# Diffusion Model Quantization: A Review

Qian Zeng, Chenggong Hu, Mingli Song, Jie Song

Zhejiang University

$\{qianz, huchenggong, songml, sjie\}$ @zju.edu.cn

Survey Project: `https://github.com/TaylorJocelyn/Diffusion-Model-Quantization`

## Abstract

Recent success of large text-to-image models has empirically underscored the exceptional performance of diffusion models in generative tasks. To facilitate their efficient deployment on resource-constrained edge devices, model quantization has emerged as a pivotal technique for both compression and acceleration. This survey offers a thorough review of the latest advancements in diffusion model quantization, encapsulating and analyzing the current state of the art in this rapidly advancing domain. First, we provide an overview of the key challenges encountered in the quantization of diffusion models, including those based on U-Net architectures and Diffusion Transformers (DiT). We then present a comprehensive taxonomy of prevalent quantization techniques, engaging in an in-depth discussion of their underlying principles. Subsequently, we perform a meticulous analysis of representative diffusion model quantization schemes from both qualitative and quantitative perspectives. From a quantitative standpoint, we rigorously benchmark a variety of methods using widely recognized datasets, delivering an extensive evaluation of the most recent and impactful research in the field. From a qualitative standpoint, we categorize and synthesize the effects of quantization errors, elucidating these impacts through both visual analysis and trajectory examination. In conclusion, we outline prospective avenues for future research, proposing novel directions for the quantization of generative models in practical applications. The list of related papers, corresponding codes, pre-trained models and comparison results are publicly available at the survey project homepage.

***Keywords*—** Diffusion models, Transformers, Model acceleration, Diffusion model quantization

# Contents

# 1 Introduction

Diffusion models [1–4] have rapidly emerged as predominant deep generative models. Through meticulous modeling of posterior probabilities and an iterative denoising framework, they facilitate precise reconstruction of sample details, significantly enhancing generation quality and fidelity. Compared to Variational Autoencoders (VAEs) [5], diffusion models exhibit superior capacity for capturing nuanced features while circumventing the inherent constraints of conventional reconstruction-based approaches in producing high-fidelity generations. Moreover, diffusion models adopt a training methodology grounded in maximum likelihood estimation, which provides a solid theoretical foundation and adeptly alleviates prevalent challenges such as mode collapse and artifactual generation—issues frequently encountered in the adversarial training paradigm of Generative Adversarial Networks (GANs) [6]. As a result, diffusion models achieve a superior balance between sample diversity and generation fidelity. Recent breakthroughs in interdisciplinary research have underscored the remarkable versatility of diffusion models across a spectrum of generative applications, including text-to-image [7, 8], image super-resolution [9, 10], inpainting [11, 12], style transfer [13–15], text-to-video [16–18], time series modeling [19, 20], interpretability modeling [21], molecule generation [22] and medical image reconstruction[23, 24].

However, diffusion models incur significant computational and memory overhead during inference. For instance, even on the high-performance A6000 GPUs, Stable Diffusion [25] with 16GB of running memory still takes over one second to perform one denoising step [26]. This inefficiency stems primarily from two critical bottlenecks: the lengthy denoising chain-often spanning up to 1000 steps [2] and the computationally demanding noise estimation network (i.e., the score estimation network [4]). To address the first bottleneck, researchers have focused on devising more efficient sampling trajectories. One line of work emphasizes the development of high-performance samplers, which accelerate the process by numerically solving the reverse-time SDE [27] or its corresponding ODE [28, 29], aiming to control discretization error while optimizing sampling step size. Another line of research explores more efficient diffusion mechanisms, such as diffusion scheme learning [30–32] and noise scale learning [33–35]. To overcome the second challenge, model compression paradigms such as pruning [36, 37], distillation [38, 39], and quantization [40, 41] have been adapted to diffusion models. Pruning reduces complexity but disrupts weight structure, often necessitates resource-intensive retraining from scratch. Distillation effectively minimizes sampling steps by learning the integral over the reverse-time SDE, yet it demands extensive data and high computational resources. For example, INSTAFLOW [32] introduces an efficient text-to-image generation model, leveraging Rectified Flow [42] as a teacher model for supervised distillation training. However, the training process still requires a significant computational cost of 199 A100 GPU days. In contrast, **Model Quantization** [43] strikes an effective balance between representational fidelity and computational efficiency, emerging as a prominent solution for acceleration in edge deployment. Consequently, it has attracted considerable attention in the development of lightweight diffusion models.

Recent breakthroughs in diffusion model quantization [40, 41, 44–46] have catalyzed a surge of cutting-edge research, predominantly focused on adapting sophisticated quantization paradigms—originally developed for CNN architectures [47] or state-of-the-art LLM techniques [48, 49]—to the unique demands of diffusion models. The seminal work PTQ4DM [40] pioneered the use of Gaussian-distributed time-step sampling for calibration set generation, establishing a crucial foundation. Subsequent innovations by Liu et al. [50] introduced enhanced distribution alignment calibration, significantly improving the representational fidelity of calibration samples. Addressing cross-time-step quantization parameter sharing from a novel perspective, So et al. [45] proposed temporal dynamic quantization, enabling step-specific activation quantization. Wang et al. [51] developed a differentiable framework for coarse-grained time-step grouping, while Huang et al. [52] introduced temporal feature maintenance quantization to mitigate sampling disorder. Tian et al. [53] further advanced temporal alignment for video generation tasks.

Quantization-aware training approaches like Q-DM [54], QuEST [55], and MEFT-QDM [56] have systematically optimized various objectives through empirical studies. The integration of LoRA techniques has pushed the boundaries of 4-bit activation quantization, with He et al. [57] proposing QaLoRA and Guo et al. [58] developing IntLoRA for fine-tuning large text-to-image models. For extreme quantization scenarios, methods including BLD [59], BinaryDM [60], and BiDM [61] have employed Bernoulli-based latent space reconstruction, while BitsFusion [62] and BDM [63] have devised mixed-precision strategies. However, these approaches often suffer from significant quantization-induced sampling perturbations, prompting the development of error-correction mechanisms like PTQD [44], D$^2$-DPM [46], and Tac-QDM [64]. In the quantization of diffusion transformers (DiT) [65], He et al. [66] and Q-DiT [67] proposed tailored group quantization mechanisms to reconcile the trade-off between quantization efficiency and performance degradation induced by outlier activations. Meanwhile, alternative approaches—including PTQ4DiT [68], DiT-AS [69], ViDiT-Q [70], and HQ-DiT [71]—alleviate quantization sensitivity through innovative channel smoothing and equalization techniques. While these contributions collectively address the predominant challenges in diffusion model quantization, many of them tend to rely on inherently repetitive strategies to solve similar issues. This underscores a critical gap in the field: the field currently lacks a comprehensive survey that systematically analyzes these issues and their corresponding solutions from a global perspective.

To advance the development of effective diffusion models, this survey offers a methodical and specialized review of prominent quantization approaches. We begin by introducing fundamental concepts in diffusion models and model quantization. In contrast to previous surveys that focused on static, one-step models, we adopt a holistic approach that considers the unique multi-step sampling dynamics inherent to diffusion processes. This allows for a comprehensive examination of quantization challenges and their underlying factors, culminating in a refined classification of solutions. We highlight pivotal concepts from each domain, empowering researchers to combine complementary strategies for optimal outcomes. Finally, we evaluate open-source quantization schemes across three well-established tasks—class-conditional, unconditional, and text-guided image generation—and offer a theoretical analysis of quantization artifacts (such as color bias, overexposure, and blurring) through both empirical and visual studies.

In summary, the contributions of this survey are outlined as follows:

- **The pioneering survey on diffusion model quantization.** To the best of our knowledge, this paper represents the first exhaustive survey of diffusion model quantization, encompassing cutting-edge advancements up to March 2025, bridging the gap between theory and practice.

- **In-depth analysis of the challenges in diffusion model quantization.** We present the first thorough synthesis of the fundamental challenges encountered in diffusion model quantization. These include quantization issues stemming from the structural characteristics of models, such as the skip connections in U-Net, attention and feedforward networks in Transformers, and the multimodal alignment mechanisms in text-to-image models, as well as challenges arising independently of model architecture, primarily due to the multi-step sampling process intrinsic to diffusion models.

- **A comprehensive taxonomy of existing approaches.** We propose an exhaustive taxonomy that categorizes the wide range of approaches in diffusion model quantization research. This includes all prominent post-training quantization (PTQ) and quantization-aware training (QAT) methods for U-Net-based diffusion models and Diffusion Transformers, accompanied by detailed explanations of the core concepts. Additionally, to cater to researchers from interdisciplinary fields, we first introduce the basic distinctions between PTQ and QAT, and subsequently classify and summarize solutions across various quantization components, including calibration sampling strategies, dynamic activation, error correction mechanisms, group partitioning, and channel equalization.

- **Quantitative benchmarking and qualitative evaluations.** We assess leading open-source solutions across three key benchmarks: class-conditional image generation, unconditional generation, and text-conditional guided image generation. Through visual analysis, we identify artifacts introduced by quantization distortions, such as color bias, pixel overexposure, detail blurring, and structured features altering, and perform a qualitative analysis supported by empirical research.

- **Frontiers and future directions:** We examine emerging and prospective challenges within the widely-adopted diffusion model framework and delineate promising avenues for future research, including the integration of quantization with advanced training strategies, the optimization of vector quantization across diverse modalities in diffusion modeling, and other insightful and forward-thinking research directions.

The remainder of this survey is structured as follows: In Sec. 2, we first present the theoretical foundations of diffusion models and model quantization, followed by an in-depth exploration of the principal challenges in diffusion model quantization. Building upon the discussions in Sec. 2, we classify and elucidate existing quantization frameworks in Sec. 3. In Sec. 4, we establish standard benchmarks and provide a thorough evaluation of prominent and open-source quantization solutions. Finally, Sec. 5 concludes the survey, highlighting several promising avenues for future research.

## 2 Background and Preliminary

### 2.1 Diffusion Models

**Discrete Space Diffusion** Sohl-Dickstein et al. [1] are the first to introduce diffusion models, initially formulated in discrete space. The model comprises two key processes: forward noise addition and reverse denoising. In the forward process, starting from clean data $\mathbf{x}_0 \sim q(\mathbf{x}_0)$, isotropic Gaussian noise is progressively added at each time step $t$ based on a predefined variance schedule $\beta_1, \ldots, \beta_T \in (0, 1)$, as defined in Eqn. (1). This step forms a Markov probability chain, detailed in Eqn.(2). As $T \to \infty$, the data distribution converges to a standard Gaussian distribution.

$$q(\mathbf{x}_t \mid \mathbf{x}_{t-1}) = \mathcal{N}(\mathbf{x}_t; \sqrt{1-\beta_t}\mathbf{x}_{t-1}, \beta_t \mathbf{I}), \tag{1}$$

$$q(\mathbf{x}_{1:T} \mid \mathbf{x}_0) = \prod_{t=1}^{T} q(\mathbf{x}_t \mid \mathbf{x}_{t-1}). \tag{2}$$

By recursively applying the reparameterization trick and introducing $\alpha_t = 1 - \beta_t$, $\bar{\alpha}_t = \prod_{i=1}^{T} \alpha_i$, it is possible to directly perturb $\mathbf{x}_0$ with noise to obtain $\mathbf{x}_t$:

$$q(\mathbf{x}_t \mid \mathbf{x}_0) = \mathcal{N}(\mathbf{x}_t; \sqrt{\bar{\alpha}_t}\mathbf{x}_0, (1-\bar{\alpha}_t)\mathbf{I}). \tag{3}$$

In the reverse process, sampling begins with $\mathbf{x}_T \sim \mathcal{N}(\mathbf{0}, \mathbf{I})$. Each preceding sample $\mathbf{x}_{t-1}$ is derived using the posterior distribution $p_\theta(\mathbf{x}_{t-1} \mid \mathbf{x}_t)$, reparameterized via a neural network within the framework of variational inference, as defined in Eqn.(4):

$$p_\theta(\mathbf{x}_{t-1} \mid \mathbf{x}_t) = \mathcal{N}(\mathbf{x}_{t-1}; \mu_\theta(\mathbf{x}_t, t), \Sigma_\theta(\mathbf{x}_t, t)). \tag{4}$$

Following the reverse Markov chain, samples are generated iteratively by applying Eqn. (4), culminating in the final synthesized image $\mathbf{x}'_0$. During training, the model parameters are learned by optimizing the variational lower bound, which effectively minimizes the negative log-likelihood [2].

**Continuous Space Diffusion** Song et al. [4] extend the discrete-time propagation chain to continuous-time space by stochastic differential equations (SDEs). In this theoretical framework, the forward diffusion process can be modeled as a solution to an Itô SDE:

$$d\mathbf{x} = \mathbf{f}(\mathbf{x}, t)\, dt + g(t)\, d\mathbf{w}, \tag{5}$$

where $\mathbf{w}$ is the standard Wiener process (*a.k.a.*, Brownian motion), $\mathbf{f}(\cdot, t) : \mathbb{R}^d \to \mathbb{R}^d$ is a vector-valued function called the drift coefficient of $\mathbf{x}(t)$, and $g(\cdot) : \mathbb{R} \to \mathbb{R}$ is a scalar function known as the diffusion coefficient of $\mathbf{x}(t)$.

In the reverse process, the predominant sampling methods are categorized into deterministic and stochastic sampling. Stochastic sampling follows Anderson's reverse-time SDE:

$$d\mathbf{x} = \left[\mathbf{f}(\mathbf{x}, t) - g(t)^2 \nabla_{\mathbf{x}} \log p_t(\mathbf{x})\right] dt + g(t) d\bar{\mathbf{w}}, \tag{6}$$

where $\nabla_{\mathbf{x}} \log p_t(\mathbf{x})$ is the score function [35], $\bar{\mathbf{w}}$ is a standard Wiener process when time flows backwards from T to 0, and dt is an infinitesimal negative timestep. Deterministic sampling follows the *probability flow* ODE, which shares the same marginal probability as the reverse-time SDE:

$$d\mathbf{x} = \left[\mathbf{f}(\mathbf{x}, t) - \frac{1}{2}g(t)^2 \nabla_{\mathbf{x}} \log p_t(\mathbf{x})\right] dt. \tag{7}$$

To estimate the score $\nabla_{\mathbf{x}} \log p_t(\mathbf{x})$ in Eqn. (6) and Eqn. (7), it is common to train a time-independent score-based model $\mathbf{s}_\theta(\mathbf{x}, t)$, which is linearly related to the noise estimation network $\epsilon_\theta$:

$$\mathbf{s}_\theta(\mathbf{x}, t) \triangleq -\frac{\epsilon_\theta(\mathbf{x}_t, t)}{\sigma_t}, \tag{8}$$

where $\sigma_t$ is the standard deviation of $p(\mathbf{x}_t|\mathbf{x}_0)$, referred to as the noise schedule.

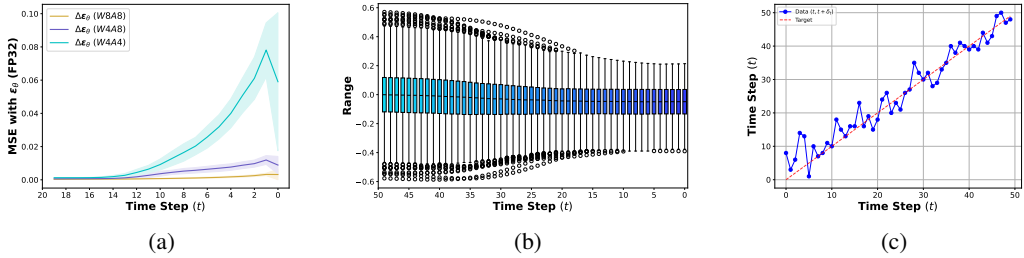

Figure 1: The quantization challenges arising from diffusion mechanisms. (a)Under W8A8, W4A8, and W4A4 quantization, the quantization noise $\Delta \epsilon_\theta$ accumulate rapidly at different rates over time steps. (b) Activation ranges of $\mathbf{x}_t$ across all 50 time steps of FP32 DDIM model on CIFAR-10. (c) Temporal feature mismatch. The coordinates of the inflection points on the blue curve can be denoted as $(t, t+\delta_t)$. This indicates that the quantized time embedding $\widehat{emb}_t$ at timestep $t$ exhibits the highest similarity with the full-precision time embedding $emb_{t+\delta_t}$ at timestep $t + \delta_t$.

## 2.2 Model Quantization

**Model quantization** [72] is a prevalent technique for model compression and acceleration, widely utilized for efficient deployment on edge devices [43]. It is primarily divided into two approaches: post-training quantization (PTQ) [43, 73, 74] and quantization-aware training (QAT) [75–77]. Fundamentally, quantization is a numerical mapping method that compresses high-precision floating-point numbers into low-bit integers. The process primarily involves three key parameters: scale factor $s$, zero point $z$, and quantization bit-width $b$. Uniform quantization is typically employed, where a floating-point value $x$ is mapped to a fixed-point value $x_{int}$ using the aforementioned parameters:

$$x_{\text{int}} = \text{clamp}\left(\left\lfloor \frac{x}{s} \right\rceil + z, 0, 2^b\right), \quad (9)$$

where $\lfloor \cdot \rceil$ is the round operation and clamp is a truncation function.

**Post-training quantization** take a pre-trained FP32 network and convert it directly into a fixed-point network without the need for the original training pipeline. This method can be data-free or require a small calibration dataset. For algorithms without any end-to-end training, the clipping thresholds $q_{min}$ and $q_{max}$, which define the scale factor $s$, are typically derived through the optimization of a local cost function $\mathcal{L}_{local}$:

$$s = \frac{q_{max} - q_{min}}{2^b}, \quad (10)$$

$$\mathcal{L}_{local} = \underset{q_{\min}, q_{\max}}{\arg\min} \; F\left(\psi(\mathbf{v}), \psi\left(\hat{\mathbf{v}}(s, z)\right)\right). \quad (11)$$

Here, $\hat{\mathbf{v}}(s, z)$ denotes the tensor quantized by applying the parameters $s$ and $z$, $F(\cdot, \cdot)$ denotes the cost function, commonly including mean squared error (MSE) or cross-entropy loss, while $\psi(\cdot)$ denotes the transformation function, typically an identity mapping for MSE and softmax for cross-entropy loss.

**Quantization-aware training** enables low-bit quantization (4 bits or fewer) by explicitly optimizing both the quantization parameters and model weights. Specifically, QAT incorporates simulated quantization to model quantization noise and employs task-specific loss functions to supervise end-to-end training. To address the non-differentiability of $\lfloor \cdot \rceil$ operation during backpropagation, QAT utilizes the Straight-Through Estimator (STE) [78] to approximate its gradients, enabling effective and efficient optimization. For further details, please refer to the quantization white paper [43].

## 2.3 Challenges in Quantizing Diffusion Models

Although seminal works such as AdaRound [79], LSQ [80] and BREQCQ [47] within traditional convolutional architectures, alongside cutting-edge advancements in the transformer-based LLM domain, including SmoothQuant [74], PreQuant [81], and PEQA [82], have robustly established the efficacy of quantization algorithms for model compression and acceleration, the application of these state-of-the-art quantization techniques to diffusion models remains a formidable challenge due to their unique multi-step iterative denoising process. Given that diffusion models can be fundamentally decomposed into two core components: the diffusion mechanism and the noise estimation model. Accordingly, the quantization challenges can be classified into two primary categories: those arising from the diffusion mechanism and those stemming from the architectural characteristics of the noise estimation model. A detailed analysis of these challenges is presented below.

### 2.3.1 Challenges from the Diffusion Mechanism

The multi-step noise elimination mechanism is a fundamental distinction between diffusion models and traditional one-step architectures like CNNs and transformers. At each time step $t$, the output $\epsilon_t$ of the noise estimation network $\epsilon_\theta$, is processed through a sampling formula to compute the next input $\mathbf{x}_{t-1}$. This iterative process introduces inherent challenges during quantization.

**C #1 Activation distributions vary across time steps.** At time step $t$, the model input $\mathbf{x}_t \sim p_\theta(\mathbf{x}_t|\mathbf{x}_{t+1})$. After inference, the output $\epsilon_t$ is transformed by the sampling formula to produce $\mathbf{x}_{t-1}$, which follows a new distribution $p_\theta(\mathbf{x}_{t-1}|\mathbf{x}_t)$. This temporal evolution of input distributions results in layer-wise activation values varying significantly across time steps, as illustrated in Fig. 1b, thereby increasing the difficulty of quantization.

**C #2 Quantization errors accumulate across time steps.** In single-step noise estimation, alignment errors introduced during quantization reconstruction propagate and accumulate across layers. These errors are further amplified by the iterative nature of diffusion models, leading to the progressive accumulation of quantization errors over sampling time steps, as shown in Fig. 1a, ultimately causing the sampling trajectory to deviate gradually.

**C #3 Temporal confusion phenomenon.** The time schedule $\{t_i\}_{i=1}^N$ in diffusion models is critical, as it uniquely maps the noise schedule $\{\sigma_i\}_{i=1}^N$, determining the distributional transition trajectory during sampling [35]. However, quantization of the TimeEmbedding layer may lead to inconsistencies in the representation of $t_{embed}$ between its pre- and post-quantization states, as shown in Fig. 1c, potentially causing disruptions in the sampling trajectory, such as retracing or convoluted, meandering transitions.

### 2.3.2 Challenges from Model Architectures

Existing diffusion models can be broadly categorized into two architectural paradigms: UNet-based diffusion and Diffusion Transformers (DiT) [65]. While both share the same underlying diffusion mechanism, the differences in the design of the noise estimation network impose distinct inherent constraints and challenges during quantization. Practically, the UNet architecture presents the following challenges:

**C #4 Bimodal data distribution from concatenate operations.** The concatenate operation in the shortcut layers of UNet directly combines shallow and deep features, inherently resulting in a bimodal distribution of weights and activations due to the significant variation in data ranges across different channels [41]. Quantization under these conditions can exacerbate uneven compression across channels, further amplifying inconsistencies in feature representation.

**C #5 Misalignment of textual and visual feature representations due to cross-attention quantization.** In UNet-based text-to-image models, such as Stable Diffusion [25], cross-attention modules inject textual information, enabling precise alignment between textual and visual features and supporting tasks like personalized generation. However, quantization may disrupt this alignment, impairing the model's ability to accurately integrate multimodal information.

Additionally, the primary challenge in quantizing DiT is:

**C #6 Data Variance Across Multiple Levels.** DiTs, adhering to the best practices of Vision Transformers (ViTs) [83], which have demonstrated superior scalability in visual recognition tasks, adopt a token-based representation of visual features for transformer-based processing. Research [84, 67] indicates that token-based visual features exhibit substantial temporal and spatial variations. In Diffusion Transformers (DiTs), significant weight disparities exist across distinct architectural components – notably global attention mechanisms, text-conditioned cross-attention modules, and temporal attention operators. These variations, compounded by timestep conditioning's systemic impact on DiT activation patterns, amplify multi-scale discrepancies in activation distributions. Such multi-level heterogeneity poses critical challenges for achieving both quantization efficiency and precision in diffusion model deployment.

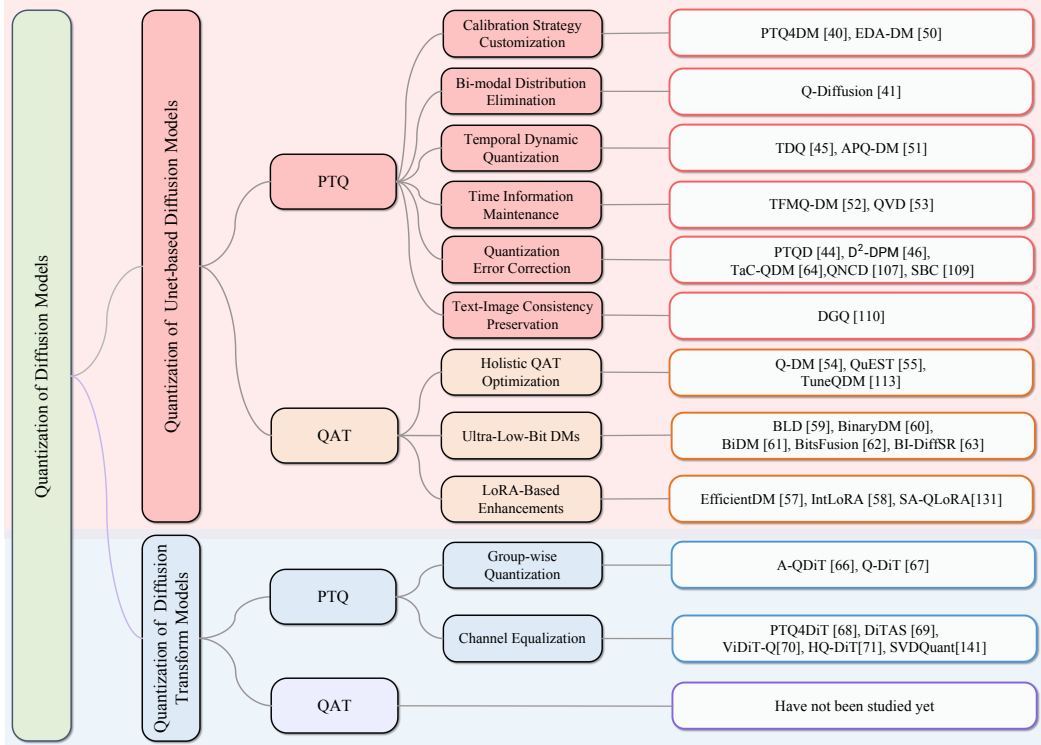

Figure 2: A comprehensive taxonomy of quantization algorithms for diffusion models, organized by quantization techniques such as PTQ (Post-Training Quantization) and QAT (Quantization-Aware Training), along with their specific methods and enhancements. The diagram highlights various strategies for optimizing quantization components, including calibration strategy customization, dynamic quantization, and low-bit methods, among others.

Existing work primarily addresses challenges within the two aforementioned categories. Building on these, this paper provides a detailed taxonomy and analysis of current quantization approaches for diffusion models.

## 3 A Taxonomy of Quantization Algorithms for Diffusion Models

This study focuses on reviewing the emerging research on efficient quantization approaches for diffusion models over the past two years, encompassing the literature available up to March 2025. We propose a comprehensive taxonomy, systematically classifying the collected works through a structured framework. Following this, we highlight representative methods within each category, particularly those addressing distinctive challenges or motivations not covered in prior studies. Specifically, for each method, we identify its foundational motivations, map them to the challenges outlined in Sec. 2.3, and provide a detailed analysis of the proposed solutions. Finally, detailed experimental evaluations conducted on a unified benchmark are presented in Sec. 4.

In detail, the proposed taxonomy is shown in Fig. 2. The complete set of algorithms is initially categorized by the architecture of the noise estimation network into Unet-based diffusion models and diffusion transformers. Each of these categories is then further divided into two types based on the quantization strategy: PTQ and QAT.

Table 1: High-level Overview of Popular Quantized Diffusion Models. The "model type" consists of two categories: I refers to U-Net-based models, and II refers to transformer-based models. MP denotes mixed precision. The mainstream benchmarks are labeled in the form of {**Dataset** (**Model**)} as follows: ❶ CIFAR10 32×32 [85] (DDIM [86]), ❷ ImageNet 64×64 [87] (DDIM), ❸ ImageNet 256×256 (LDM-4 [25]), ❹ LSUN-Bedrooms 256×256 [88] (LDM-4), ❺ LSUN-Churches 256×256 [88] (LDM-8 [25]), ❻ CelebA-HQ 256×256 [89] (LDM-4), ❼ FFHQ 256×256 [90] (LDM-4), ❽ MS-COCO [91] (Stable Diffusion v1-4 or XL[25]), ❾ Stable-Diffusion-Prompts (Stable Diffusion v1-4 or XL), ❿ PartiPrompts [92] (Stable Diffusion v1-4, v1-5 or XL), ⓫ FS-COCO [93] (MagicAnimate [94]) + MS-COCO (AnimateDiff [95]), ⓬ OPT[96] (LAION-5B[97]), ⓭ ImageNet 256×256 and 512×512 (DiT-XL/2 [65]), ⓮ MS-COCO 512×512 (conditional DiT [65]), ⓯ MS-COCO 512×512 (PixArt[98]) + Open-Sora (STDiT [99]), ⓰ DreamBooth bench [100], ⓱ DreamBooth, ADE20K [101] and CelebA-HQ dataset (Stable Diffusion v1-5) + StyleDrop dataset [102] (Stable Diffusion XL), ⓲ Set5 [103] + B100 [104] + Urban100 [105] + Manga109 [106].

| | Method | Model Type | Bitwidth | Benchmark | Open Source | Highlight |
|---|---|---|---|---|---|---|
| PTQ-based Approach | PTQ4DM [40] | I | W8A8, W4A8 | ❶❷ | ✓ | Generating sample sets from the denoising pipeline is proven optimal, leading to the proposal of the Normally Distributed Time-step Calibration Collection (DNTC) algorithm. |
| | Q-Diffusion [41] | I | W8A8, W4A8 | ❶❹❺❻ | ✓ | The Split operation reduces the impact of the inherent bimodal data distribution in U-Net on quantization, and introduces a calibration method based on uniform time-step sampling. |
| | EDA-DM [50] | I | W8A8, W4A8 | ❶❸❹❺❻ | ✓ | A new calibration strategy is proposed to improve distribution alignment, complemented by fine-grained reconstruction to address shallow underfitting and deep overfitting in block reconstruction. |
| | TFMQ-DM [52] | I | W8A8, W4A8 | ❸❹❺❻❼❽ | ✓ | An independent time information alignment module is introduced to explicitly enhance temporal consistency during quantization. |
| | QVD [53] | I | W8A8, W4A8 | ⓫ | ✓ | Applying time feature alignment to the quantization task of video generation models. |
| | APQ-DM [51] | I | W8A8, W6A6 | ❶❸❹❺❻ | ✓ | Introducing a differentiable group assignment strategy enables distribution-aware dynamic quantization, with a calibration set selected based on the principle of structural risk minimization. |
| | PTQD [44] | I | W8A8, W4A8 | ❸❹❺ | ✓ | Quantization noise is deconstructed into linearly correlated and uncorrelated components with respect to the full-precision output, with error correction performed for each component separately. |
| | QNCD [107] | I | W8A8, W4A8, W4A6 | ❶❸❹❽ | ✓ | A linear smoothing factor is introduced to mitigate the impact of large variance between channels, along with a quantization noise estimation mechanism to filter it. |
| | D²-DPM [46] | I | W8A8, W4A8 | ❸❹❺ | ✓ | A joint probability model of quantized output and quantization noise is first established, from which a conditional Gaussian model for quantization noise is derived. Two variants of a dual denoising mechanism are then proposed based on sampling equation optimization. |
| | TaC-QDM [64] | I | W8A8, W4A8, W3A8, W2A8 | ❶❹❺ | ✗ | Input bias correction and quantization noise reconstruction are introduced to perform error correction on $x_t$ and $\epsilon_\theta^{(t)}$, respectively. |
| | TMPQ-DM [108] | I | W6A6, W7A7, MP | ❶❸❹❺❻ | ✗ | The timestep sub-sequence search and layer-wise quantization precision are jointly optimized within a unified search space using a gradient-free evolutionary search algorithm. |
| | TDQ [45] | I | W8A8, W4A8, W4A4, W3A3 | ❶❺ | ✗ | Train a multilayer perceptron to dynamically predict the quantization parameters for each timestep's activation. |
| | SBC [109] | I | FP32-SM8, W8A16-SM8 | ❽ | ✗ | An offline softmax bias correction strategy is proposed. |
| | DGQ [110] | I | W8A8, W4A8, W4A6 | ❽❿ | ✓ | Distributed-aware group quantization is applied to address quantization challenges caused by outliers and specific structures in cross-attention. |
| | PCR [111] | I | W8A8, W4A8 | ❽❿ | ✓ | A benchmark for text-to-image quantization, QDiffBench, is proposed; progressive activation calibration is introduced, considering the accumulation of activation quantization errors across steps. |
| | MixDQ [112] | I | W8A8, W4A8 | ❽ | ✗ | After analyzing the sensitivity of several modules to quantization, an optimal mixed-precision configuration scheme is derived using integer programming. |
| | Q-DiT [67] | II | W8A8, W4A8 | ⓭ | ✓ | Dynamic group quantization is applied to mitigate the impact of structured outliers in activation channels within DiT. |
| | PTQ4DiT [68] | II | W8A8, W4A8 | ⓭ | ✓ | Channel rebalancing is applied to reduce the significant data variance between channels. |
| | DiTAS [69] | II | W8A8, W4A8 | ⓭ | ✓ | Temporal-aggregated smoothing techniques are applied to mitigate the impact of channel-wise outliers within the input activations. |
| | A-QDiT [66] | II | W8A8, W4A8 | ⓮ | ✗ | The superiority of 1-step calibration over multi-step calibration in DiT quantization is validated, and weight group quantization alleviates poor quantization due to weight dispersion. |
| | ViDiT-Q [70] | I | W8A8, W4A8, W6A6 | ⓯ | ✓ | Applying fine-grained grouping dynamic quantization with static-dynamic channel balancing for large scale video DiTs. |
| | HQ-DiT [71] | I | W8A8, W4A8, W4A4 | ❸ | ✗ | A first instance where both weights and activations in DiTs are quantized to just 4 bits using floating-point precision. |
| QAT-based Approach | Q-DM [54] | I | W8A8, W4A4, W3A3, W2A2 | ❶❷ | ✗ | The quantization function for each timestep is redefined by incorporating statistics from multiple timesteps, and a QAT optimization objective based on Noise-estimating Mimicking is designed. |
| | QuEST [55] | I | W8A8, W4A8, W4A4 | ❸❹❺❻ | ✓ | It demonstrates the necessity of fine-tuning weight parameters under low-bit quantization, analyzes the quantization-sensitive layers, and proposes a corresponding joint optimization objective. |
| | TuneQDM [113] | I | W8A32, W4A32 | ⓰ | ✗ | The PEQA baseline is optimized, enabling scale updates to focus on multiple channels. |
| | EfficientDM [57] | I | W8A8, W4A8, W4A4, W2A8 | ❶❸❹❺ | ✓ | A data-free, quantization-aware, and parameter-efficient fine-tuning framework for low-bit diffusion models is introduced. |
| | IntLoRA [58] | I | W8A32, W4A32 | ⓱ | ✓ | By setting up adaptation-quantization separation and multiplicative low-rank adaptation, a fully integer-based LoRA-PEFT process is implemented, which maintains both the pre-trained weights and LoRA weights as integer values, thereby reducing the fine-tuning overhead. |
| | BLD [59] | I | W2A2 | ❸❹❺❻❼ | ✓ | It models the bi-directional mappings between an image and its corresponding latent binary representation by training an auto-encoder with a Bernoulli encoding distribution. |
| | BinaryDM [60] | I | W1A8, W1A4 | ❶❸❹❻❼ | ✓ | The Evolvable-Basis Binarizer (EBB) is proposed to optimize binary representation, and a Low-rank Representation Mimicking (LRM) objective is designed to assist in the optimization of binarized diffusion models. |
| | BiDM [61] | I | W1A1 | ❶❹❺❼ | ✓ | It uses learnable activation binarizers and cross-timestep feature connections to address the highly timestep-correlated activation features of DMs and proposes Space Patched Distillation (SPD) to tackle the difficulty of matching binary features during distillation. |
| | BitsFusion [62] | I | W1.99A32 | ❿ | ✗ | A specialized low-bit diffusion model initialization and a Two-Stage Training Pipeline are proposed. |
| | BI-DiffSR [63] | I | W2A2 | ⓲ | ✓ | A binary-friendly structure is added to the original Unet, and a Timestep-aware Redistribution/Activation Function is set up to accommodate dynamically changing activations over time. |

## 3.1 Unet-Based Diffusion Model Quantization

### 3.1.1 PTQ-Based Approaches

As derived from Eqn. (11), the most basic PTQ optimize a local cost function to determine the quantization scale factor $s$ and zero-point $z$. For weight quantization, given the pretrained weights $w$, the optimal parameter set $(s_{opt}, z_{opt})$ can be obtained through a heuristic iterative search initialized with $(s_{init}, z_{init})$. However, activation quantization typically requires sampling a calibration dataset to propagate through the network and capture intermediate activation outputs for computing the activation quantization parameters.

Building on this foundation, BRECQ [47] introduces a block-wise reconstruction mechanism that further refines the PTQ process, establishing itself as a widely adopted baseline for post-training quantization of diffusion models [40, 41, 44]. Specifically, it first computes the weight quantization parameters $(s_w, z_w)$ based on Eqn. (11). Then, it substitutes the rounding operator in Eqn. (9) with an adaptive rounding function [79] parameterized by a learnable variable $\alpha$:

$$Sigmoid(\alpha) = \frac{x}{s_w} - \lfloor \frac{x}{s_w} \rfloor, \tag{12}$$

$$R_\alpha(x, s_w) = \lfloor \frac{x}{s_w} \rfloor + \mathbb{I}(\alpha \geq 0), \tag{13}$$

$$x_{\text{int}} = \text{clamp}\left(R_\alpha(x, s_w) + z_w, 0, 2^b\right), \tag{14}$$

where $\lfloor \cdot \rfloor$ is the floor function and $\mathbb{I}$ is the indicator function. When quantizing the module $f_i(\cdot, \cdot)$, calibration samples $\{(\mathbf{x}_{t_i}, t_i)\}_{i=1}^N$ are passed through the model to capture the output activations $f_i(\mathbf{x}_{t_i}, t_i)$ and $\hat{f}_i(\mathbf{x}_{t_i}, t_i)$, corresponding to the activations before and after quantization, respectively. The mean squared error (MSE) between $f_i(\mathbf{x}_{t_i}, t_i)$ and $\hat{f}_i(\mathbf{x}_{t_i}, t_i)$ is subsequently employed as the reconstruction loss $\mathcal{L}_{recon}$ to optimize the adaptive parameter $\alpha$ and the activation quantization parameter $s_a$:

$$\mathcal{L}_{recon} = \arg\min_{\alpha, s_a} \|f_i(\mathbf{x}_{t_i}, t_i) - \hat{f}_i(\mathbf{x}_{t_i}, t_i)\|^2. \tag{15}$$

Unlike QAT, which directly optimizes the quantization parameters for the entire model, BRECQ employs module-wise reconstruction. Additionally, Eqn. (12) constrains the optimization space of $\alpha$, ensuring a more efficient optimization process. Recent advancements studies build upon this algorithm, incorporating plug-and-play components such as calibration dataset sampling strategies and alignment objectives tailored to the diffusion framework, with the goal of enhancing performance in downstream tasks.

**Calibration Strategy Customization.** We start with the pioneering work PTQ4DM [40] proposed by Shang et al., which identifies the primary challenge as **C #1**. Innovatively, this work is the first to propose a tailored calibration dataset designed to precisely capture the input distribution of the network $\epsilon_\theta$ across multiple timesteps, effectively addressing the challenge of activation range variations in multi-timestep scenarios. To this end, Shang et al. conducted extensive empirical studies, initially evaluating three fundamental sampling strategies: sampling from the diffusion starting point $\mathbf{x}_0$, the noisy inputs $\mathbf{x}_t$ during forward diffusion, and the reverse diffusion starting point $\hat{\mathbf{x}}_T$. Among these, the dataset based on $\hat{\mathbf{x}}_T$ demonstrated the best quantization performance. Consequently, they prioritized leveraging the reverse sampling trajectory to construct a more effective calibration datas. Ultimately, building on the key observation that the calibration effectiveness significantly improves as $\mathbf{x}_t$ approaches $\mathbf{x}_0$, they propose the **N**ormally **D**istributed **T**ime-step **C**alibration (NDTC) method, which generates calibration samples by sampling the time schedule $\{t_i\}_{i=1}^N$ as follows:

$$t_i \sim \mathcal{N}(\mu, \frac{T}{2}) \ (i = 1, 2, \ldots, N), \tag{16}$$

where $\mu \leq \frac{T}{2}$, ensuring that $t_i$ is more densely distributed around $t_0$.

This work establishes the feasibility of data-free quantization in diffusion models by leveraging generated samples from the reverse sampling process for calibration. While formal theoretical proof is absent, NDTC's effectiveness is empirically validated, enabling seamless integration into various quantization methods.

Following the same foundational principle of aligning the distribution of the calibration set with the original distribution to the fullest extent possible, Liu et al. [114] propose a **T**emporal **D**istribution

Alignment Calibration (TDAC) method to further enhance coverage. The core idea is to evaluate the irreplaceability of the sample $\mathbf{x}_t$ at each timestep $t$ to determine its calibration sampling density. In detail, they adopt the U-Net's middle-stage output feature map $F_t$ at timestep $t$ as the representation carrier for $\mathbf{x}_t$, with the Density score $D_t$ and Variety score $V_t$ for each timestep $t$ defined as follows:

$$D_t = |\{F_i \mid \mathrm{mse}(F_t, F_i) < \epsilon, i \in [1, T]\}|, \tag{17}$$

$$V_t = \sum_{i=1}^{T} (1 - \mathrm{dist}(F_t, F_i)). \tag{18}$$

Subsequently, min-max normalization is applied to obtain the normalized scores $\hat{D}_t$ and $\hat{V}_t$, which are then used to compute the calibration sample count $N_t$ at timestep $t$ as follows:

$$S_t = \widehat{D_t} + \lambda \cdot \widehat{V_t}, \tag{19}$$

$$N_t = \frac{S_t}{\sum_{t=1}^{T} S_t} \cdot N. \tag{20}$$

Here, $\gamma$ is a tunable balancing factor, and $N$ denotes the total number of calibration samples.

The efficacy of these calibration methods is substantiated by experimental evidence. However, they tackle only a limited aspect of the broader quantization problem, leaving considerable challenges for future research.

**Bi-modal Distribution Elimination.** Concurrently, Li et al. [41] also investigate the challenge of activation range variations across multiple timesteps (**C #1**), while further pinpointing the quantization difficulties arising from the U-Net architecture (**C #4**). Similarly, they address **C #1** from the perspective of designing a calibration strategy. Through observation, they found that the output activation distributions across consecutive timesteps are highly similar and propose randomly sampling intermediate inputs uniformly at fixed intervals across all timesteps to generate a small calibration set, equivalent to defining a time schedule $\{t_i\}_{i=1}^{N}$ with $\Delta t_i = t_i - t_{i-1} = C$, where $C$ is the sampling interval. To further address the significant errors arising from quantization following the bimodal data distribution caused by shortcut layers, Li et al. propose a "split" quantization strategy. This method performs quantization prior to concatenation, applicable to both weight and activation quantization, and is mathematically expressed as follows:

$$\mathcal{Q}_X(X) = \mathcal{Q}_{X_1}(X_1) \oplus \mathcal{Q}_{X_2}(X_2), \tag{21}$$

$$\mathcal{Q}_W(W) = \mathcal{Q}_{W_1}(W_1) \oplus \mathcal{Q}_{W_2}(W_2), \tag{22}$$

$$\mathcal{Q}_X(X)\mathcal{Q}_W(W) = \mathcal{Q}_{X_1}(X_1)\mathcal{Q}_{W_1}(W_1)$$
$$+ \mathcal{Q}_{X_2}(X_2)\mathcal{Q}_{W_2}(W_2), \tag{23}$$

where $X \in \mathbb{R}^{w \times h \times c_{in}}$ and $W \in \mathbb{R}^{c_{in} \times c_{out}}$ are the input activation and layer weight, which can be naturally split into $X_1 \in \mathbb{R}^{w \times h \times c_1}$, $X_2 \in \mathbb{R}^{w \times h \times c_2}$, $W_1 \in \mathbb{R}^{c_1 \times c_{out}}$, and $W_2 \in \mathbb{R}^{c_2 \times c_{out}}$, respectively. $c_1$ and $c_2$ are determined by the concatenation operation. $\mathcal{Q}(\cdot)$ denotes the quantization operator and $\oplus$ denotes the concatenation operator. This method partially alleviates the intrinsic quantization challenges caused by the U-Net architecture and demonstrates strong transferability for quantizing UNet-based models.

**Temporal Dynamic Quantization.** The aforementioned works aim to address the challenges inherent in shared activation quantization functions by constructing calibration datasets that achieve superior coverage of varying activation distributions across multiple timesteps. This approach, however, imposes higher demands on the convergence performance of baseline reconstruction algorithms. In contrast, So et al. [45] are the first to introduce **T**emporal **D**ynamic **Q**uantization (TDQ), a method that learns a distinct set of activation quantization parameters for each timestep, thereby fundamentally circumventing the limitations associated with shared activation quantization parameters.

The TDQ module is attached to each quantization operator, with its internal parameters optimized through PTQ or QAT processes while incurring no additional inference overhead. Specifically, the module consists of an encoder $enc(\cdot)$ and a shallow multilayer perceptron $f(\cdot)$. Given an arbitrary timestep $t$ as input, it is first transformed using geometric Fourier encoding [115, 116] to mitigate the low-frequency inductive bias inherent in neural networks [117], and is then passed through $f(\cdot)$ to produce the activation quantization parameter $s_a$. This process is formalized as:

$$I = enc(t) = (sin(\frac{t}{t_{max}^{0/d}}), cos(\frac{t}{t_{max}^{0/d}}),$$
$$sin(\frac{t}{t_{max}^{2/d}}), cos(\frac{t}{t_{max}^{2/d}}), \ldots, sin(\frac{t}{t_{max}^{d/d}}), cos(\frac{t}{t_{max}^{d/d}})), \quad (24)$$

$$\tilde{s}_a = f(I). \quad (25)$$

Once the optimization process is complete, the parameters of $f(\cdot)$ are fixed, enabling the corresponding $s_a$ for each timestep $t$ to be precomputed before inference, effectively eliminating any additional computational overhead.

While this temporal dynamic quantization introduces some additional computational overhead during the quantization reconstruction process, it is inherently well-suited to the distinctive multi-timestep characteristics of diffusion model quantization and incurs no additional inference overhead. Furthermore, with the support of sampling acceleration theory [28, 118, 119], the memory footprint of the additional quantization parameters becomes negligible, making it widely integrated into subsequent works [84, 51, 120]. Among these, the PTQ-based APQ-DM [51], further transforms the problem into a grouping matching task. It partitions the time schedule $\{t_i\}_{i=1}^N$ into several groups and devises a differentiable group assignment strategy to allocate the optimal activation quantization function to each group. Finally, it optimizes the process using representative calibration samples generated based on the principle of structural risk minimization.

**Time Information Maintenance.** Nevertheless, the studies outlined above primarily concentrate on reconstructing the final activation outputs of target modules, often overlooking the shallow-layer information components. Notably, Huang et al. [52] identified the critical challenge (**C#3**), highlighting that after the quantization of the target module, the temporal information $\widehat{emb}_{t,i}$, which is crucial for reverse diffusion sampling, becomes inconsistent with its original form $emb_{t,i}$ for a given timestep $t$ and block $i$. Specifically, the temporal feature $emb_{t,i}$ is mathematically defined as:

$$emb_{t,i} = g_i(h(t)), \quad (26)$$

where $h(\cdot)$ denotes the *time embed* and $g_i(\cdot)$ represents the *embedding layer* of the $i$-th Residual Bottleneck Block. This temporal information mismatch causes a deviation in the corresponding temporal position of the image within the denoising trajectory, formulated as:

$$t \leftarrow emb_{t,i}, \ t \not\leftarrow \widehat{emb}_{t,i} \quad \Rightarrow \quad \mathbf{x}_t \not\Rightarrow \mathbf{x}_{t-1}. \quad (27)$$

Furthermore, as illustrated in Fig. 1c, the quantized temporal feature $\widehat{emb}_{t,i}$ deviates from its full-precision counterpart $emb_{t,i}$ and tends to align more closely with the inaccurate temporal feature $emb_{t+\delta_t,i}$ corresponding to $t + \delta_t$.

To address the problem of temporal feature disturbance, Huang et al. [52] first consolidate all *embedding layers* and *time embed* into a unified Temporal Information Block, denoted as $\{g_i(h(\cdot))\}_{i=0,\ldots,n}$ to facilitate a meticulous separation of the reconstruction and calibration processes for each embedding layer and Residual Bottleneck Block. Building upon this, they introduce **T**emporal **I**nformation **A**ware **R**econstruction (TIAR) to effectively minimize feature perturbations:

$$\mathcal{L}_{TIAR} = \sum_{i=0}^{n} \left\| g_i(h(t)) - \widehat{g}_i(\widehat{h}(t)) \right\|_F^2, \quad (28)$$

where $\widehat{h}(\cdot)$ and $\widehat{g}_i(\cdot)$ are quantized versions of $h(\cdot)$ and $g_i(\cdot)$.

This work demonstrates that incorporating additional temporal feature alignment enhances the overall convergence of the quantization algorithm. In a similar vein, Tian et al. [53] recognize the significance of preserving temporal features in video diffusion models (**C#3**), but emphasize enhancing their discriminability from an alternative perspective. Following the definition of time encoding layer $h(\cdot)$ and embedding function $f(\cdot)$, the temporal feature at timestep $t$ is formulated as:

$$\mathbf{T}_{emb}^t = f(h(t)). \quad (29)$$

Crucially, they further introduce the Temporal Discriminability Score (TDScore) to assess the distinctiveness of $\mathbf{T}_{emb}^t$ across different timesteps $t$. The computation comprises two steps: first, applying a logarithmic function to enhance the focus on minor values within $\mathbf{T}_{emb}^t$; second, calculating the mean cosine similarity between the feature and its $n$-contiguous timesteps:

$$\mathbf{T}_{emb}^{t'} = sign(\mathbf{T}_{emb}^t) \cdot |log_2|\mathbf{T}_{emb}^t||, \tag{30}$$

$$TDScore_t = \frac{1}{n} \sum_{i=t+1}^{i=t+n} cos\_sim\left(\mathbf{T}_{emb}^{t'}, \mathbf{T}_{emb}^{i'}\right). \tag{31}$$

Ultimately, leveraging the insight that a lower TDScore indicates higher discriminability of the temporal feature, Li et al. optimize a localized cost function $\mathcal{L}_{local}$ to determine the quantization parameters:

$$\mathcal{L}_{local} = \sum_{i=1}^{T} TDScore_i + \sum_{i=1}^{T} \left(T_{emb}^i - \widehat{T}_{emb}^i\right)^2. \tag{32}$$

Additionally, some works have further improved quantization fidelity by optimizing mixed-precision quantization schemes. These methods typically involve a more granular analysis of the quantization reconstruction process, identifying the quantization sensitivity of different modules [121, 108] or time steps [111] to allocate bit widths accordingly. However, their practicality is constrained by poor hardware compatibility.

**Text-Image Consistency Preservation.** The multimodal interaction mechanism aligns semantics by mapping text and image features into a shared embedding space, which allows for the application of control constraints. In Stable Diffusion [25], this alignment is achieved through cross-modal attention. As a result, precise attention quantization plays a crucial role in maintaining text-image consistency. Tang et al. [111] are the first to empirically demonstrate that activation near the $\mathbf{x}_T$ has a greater impact on text-image consistency, while activation near the $\mathbf{x}_0$ primarily affects image quality. Building on this, Ryu et al. [110] further investigated the patterns of activation and key attention scores to mitigate **C#5**. First, they confirm the importance of activation outliers. To safeguard these outliers, they assess the variability of activation values within each dimension and use this metric to partition the groups for quantization, thereby balancing the precision gains from preserving the outliers with the resulting quantization error. Secondly, they identify a specific attention pattern: the cross-attention scores corresponding to the $<start>$ token exhibit a distinct peak, and background pixels tend to have high attention scores (close to 1.0) for the $<start>$ token. To address this, they first isolate the $<start>$ token and, to better preserve the attention scores associated with the image content (non-background regions), which are relatively smaller, they employ logarithmic quantization.

**Quantization Error Correction.** Thus far, the works discussed above have comprehensively covered the optimization of various components in PTQ. Nevertheless, these approaches inevitably introduce small amounts of quantization noise, which accumulates across multiple timesteps (**C#2**). Consequently, the latest research [44, 46, 107, 109] has focused on leveraging multi-step denoising mechanisms to mitigate this effect.

He et al. [44] first establish a linear relationship between quantization noise $\Delta\epsilon_\theta(\mathbf{x}_t, t)$ and the full-precision output $\epsilon_\theta(\mathbf{x}_t, t)$, disentangling the quantization noise into two components:

$$\Delta\epsilon_\theta(\mathbf{x}_t, t) = k\,\epsilon_\theta(\mathbf{x}_t, t) + \Delta\epsilon_\theta^{'}(\mathbf{x}_t, t), \tag{33}$$

where $k$ is the linear correlation coefficient computed through statistical linear regression, and $\Delta\epsilon_\theta^{'}(\mathbf{x}_t, t)$ is the residual quantization noise, assumed uncorrelated with $\epsilon_\theta(\mathbf{x}_t, t)$. The sampling formula is then modified to separately correct the linearly correlated and uncorrelated components. In the case of DDPM sampling, the linearly correlated quantization noise is corrected by directly dividing the quantized output $\hat{\epsilon}_\theta(\mathbf{x}_t, t)$ by $1+k$, as demonstrated below:

$$\begin{aligned}
\mathbf{x}_{t-1} &= \frac{1}{\sqrt{\alpha_t}}\left(\mathbf{x}_t - \frac{\beta_t}{\sqrt{1-\bar{\alpha}_t}}\frac{\hat{\epsilon}_\theta(\mathbf{x}_t, t)}{1+k}\right) + \sigma_t \mathbf{z}_t \\
&= \frac{1}{\sqrt{\alpha_t}}\left(\mathbf{x}_t - \frac{\beta_t}{\sqrt{1-\bar{\alpha}_t}}\frac{\epsilon_\theta(\mathbf{x}_t, t) + \Delta\epsilon_\theta(\mathbf{x}_t, t)}{1+k}\right) + \sigma_t \mathbf{z}_t \\
&= \frac{1}{\sqrt{\alpha_t}}\left(\mathbf{x}_t - \frac{\beta_t}{\sqrt{1-\bar{\alpha}_t}}\epsilon_\theta(\mathbf{x}_t, t)\right) + \sigma_t \mathbf{z}_t \\
&\quad - \frac{\beta_t}{\sqrt{\alpha_t}\sqrt{1-\bar{\alpha}_t}(1+k)}\Delta\epsilon'_\theta(\mathbf{x}_t, t).
\end{aligned} \quad (34)$$

Subsequently, the uncorrelated noise is modeled as a Gaussian distribution through statistical parameter estimation:

$$\Delta\epsilon'_\theta(\mathbf{x}_t, t) \sim \mathcal{N}(\boldsymbol{\mu_q}, \sigma_q \boldsymbol{I}), \quad (35)$$

and updates the scheduled sampling variance to $\sigma'_t$, calculated as:

$$\sigma_t'^2 + \frac{\beta_t^2}{\alpha_t(1-\bar{\alpha}_t)(1+k)^2}\sigma_q^2 = \sigma_t^2, \quad (36)$$

$$\sigma_t'^2 = \begin{cases} \sigma_t^2 - \frac{\beta_t^2}{\alpha_t(1-\bar{\alpha}_t)(1+k)^2}\sigma_q^2, & \text{if } \sigma_t^2 \geq \frac{\beta_t^2}{\alpha_t(1-\bar{\alpha}_t)(1+k)^2}\sigma_q^2 \\ 0, & \text{otherwise.} \end{cases} \quad (37)$$

This work presents an effective approach for quantization error correction, but assumes a strong linear correlation between quantization noise and full-precision output. When this assumption breaks, the reliability of the regression-derived $k$ is compromised.

More generally, Zeng et al. [46] propose D$^2$-DPM, which decomposes the impact of quantization noise into mean bias and variance bias. The mean bias shifts the drift coefficient in the sampling equation, altering the trajectory's direction, while the variance bias modifies the diffusion coefficient, affecting the trajectory's convergence.

For the sake of clarity, we use $\hat{\boldsymbol{\mathcal{E}}}$ and $\Delta\boldsymbol{\mathcal{E}}$ to denote the variables of $\hat{\epsilon}_\theta(\mathbf{x}_t, t)$ and $\Delta\epsilon_\theta(\mathbf{x}_t, t)$, respectively. With the observed Gaussian properties of quantization noise, they introduce a time-step-aware quantization noise modeling approach, beginning with the joint modeling of $\hat{\boldsymbol{\mathcal{E}}}$ and $\Delta\boldsymbol{\mathcal{E}}$:

$$\begin{bmatrix} \hat{\boldsymbol{\mathcal{E}}} \\ \Delta\boldsymbol{\mathcal{E}} \end{bmatrix} \sim \mathcal{N}\left(\begin{bmatrix} \hat{\boldsymbol{\mu}} \\ \Delta\boldsymbol{\mu} \end{bmatrix}, \begin{bmatrix} \boldsymbol{\Sigma}_{\hat{\boldsymbol{\mathcal{E}}},\hat{\boldsymbol{\mathcal{E}}}} & \boldsymbol{\Sigma}_{\hat{\boldsymbol{\mathcal{E}}},\Delta\boldsymbol{\mathcal{E}}} \\ \boldsymbol{\Sigma}_{\Delta\boldsymbol{\mathcal{E}},\hat{\boldsymbol{\mathcal{E}}}} & \boldsymbol{\Sigma}_{\Delta\boldsymbol{\mathcal{E}},\Delta\boldsymbol{\mathcal{E}}} \end{bmatrix}\right). \quad (38)$$

Next, the distribution of the quantization noise $\Delta\boldsymbol{\mathcal{E}}$, conditioned on the quantized output $\hat{\boldsymbol{\mathcal{E}}} = \hat{\epsilon}$, is derived from the joint distribution as follows:

$$\left\{\Delta\boldsymbol{\mathcal{E}}|\hat{\boldsymbol{\mathcal{E}}} = \hat{\epsilon}\right\} \sim \mathcal{N}\left(\boldsymbol{\mu}_{\Delta\boldsymbol{\mathcal{E}}|\hat{\boldsymbol{\mathcal{E}}}=\hat{\epsilon}}, \boldsymbol{\Sigma}_{\Delta\boldsymbol{\mathcal{E}}|\hat{\boldsymbol{\mathcal{E}}}=\hat{\epsilon}}\right), \quad (39)$$

$$\boldsymbol{\mu}_{\Delta\boldsymbol{\mathcal{E}}|\hat{\boldsymbol{\mathcal{E}}}=\hat{\epsilon}} = \boldsymbol{\Sigma}_{\Delta\boldsymbol{\mathcal{E}},\hat{\boldsymbol{\mathcal{E}}}}\boldsymbol{\Sigma}_{\hat{\boldsymbol{\mathcal{E}}},\hat{\boldsymbol{\mathcal{E}}}}^{-1}(\hat{\epsilon} - \hat{\boldsymbol{\mu}}) + \Delta\boldsymbol{\mu}, \quad (40)$$

$$\boldsymbol{\Sigma}_{\Delta\boldsymbol{\mathcal{E}}|\hat{\boldsymbol{\mathcal{E}}}=\hat{\epsilon}} = \boldsymbol{\Sigma}_{\Delta\boldsymbol{\mathcal{E}},\Delta\boldsymbol{\mathcal{E}}} - \boldsymbol{\Sigma}_{\Delta\boldsymbol{\mathcal{E}},\hat{\boldsymbol{\mathcal{E}}}}\boldsymbol{\Sigma}_{\hat{\boldsymbol{\mathcal{E}}},\hat{\boldsymbol{\mathcal{E}}}}^{-1}\boldsymbol{\Sigma}_{\hat{\boldsymbol{\mathcal{E}}},\Delta\boldsymbol{\mathcal{E}}}. \quad (41)$$

To address the intractability of Eqn. (38) in high dimensions, it is assumed that the elements of $\hat{\boldsymbol{\mathcal{E}}}$ and $\Delta\boldsymbol{\mathcal{E}}$ are uncorrelated, except that corresponding elements ($i$-th) in $\hat{\boldsymbol{\mathcal{E}}}$ and $\Delta\boldsymbol{\mathcal{E}}$ can be correlated. Under this assumption, the covariance matrices $\boldsymbol{\Sigma}_{\hat{\boldsymbol{\mathcal{E}}},\hat{\boldsymbol{\mathcal{E}}}}, \boldsymbol{\Sigma}_{\hat{\boldsymbol{\mathcal{E}}},\Delta\boldsymbol{\mathcal{E}}}, \boldsymbol{\Sigma}_{\Delta\boldsymbol{\mathcal{E}},\hat{\boldsymbol{\mathcal{E}}}}$ and $\boldsymbol{\Sigma}_{\Delta\boldsymbol{\mathcal{E}},\Delta\boldsymbol{\mathcal{E}}}$ become diagonal. Furthermore, assuming isotropic distributions for $\hat{\boldsymbol{\mathcal{E}}}$ and $\Delta\boldsymbol{\mathcal{E}}$ (i.e., $\boldsymbol{\Sigma} = \sigma^2 \boldsymbol{I}$) simplifies the estimation of the joint distribution. Based on the reconstructed quantization noise $\Delta\tilde{\epsilon}(\mathbf{x}_t, t) \sim \mathcal{N}(\Delta\boldsymbol{\mu}, \sigma_\Delta^2 \boldsymbol{I})$, two variants of the dual denoising mechanism are introduced: *stochastic* dual denoising (S-D$^2$) and *deterministic* dual denoising (D-D$^2$). In S-D$^2$, the diffusion noise distribution is restored by directly subtracting the estimated quantization noise:

$$d\mathbf{x} = \left[\mathbf{f}(\mathbf{x}, t) + g(t)^2 \frac{\hat{\epsilon}_\theta(\mathbf{x}_t, t) - \Delta\epsilon'(\mathbf{x}_t, t)}{\sigma_t}\right]dt + g(t)d\bar{\mathbf{w}}. \quad (42)$$

In D-D$^2$, the distribution is restored by subtracting the conditional mean and adjusting the diffusion coefficient with the conditional variance:

$$\mathrm{d}\mathbf{x} = \left[\mathbf{f}(\mathbf{x},t) + g(t)^2 \frac{\hat{\boldsymbol{\epsilon}}_\theta(\mathbf{x}_t,t) - \Delta\boldsymbol{\mu}}{\sigma_t}\right]\mathrm{d}t + \sqrt{g^2(t) - \frac{g^4(t)\sigma_\Delta^2(t)}{\sigma_t^2}}\mathrm{d}\bar{\mathbf{w}}. \tag{43}$$

Similarly, leveraging the dynamic error correction framework to address Challenge **C#2**, Yao et al. [64] attribute the quantization error to discrepancies in noise estimation and input alignment, which are systematically analyzed through the lens of the DDPM sampling formula as follows:

$$\begin{aligned}
\|\Delta\mathbf{x}_{t-1}\| &= \|\hat{\mathbf{x}}_{t-1} - \mathbf{x}_{t-1}\| \\
&= \|\frac{1}{\sqrt{\alpha_t}}(\hat{\mathbf{x}}_t - \mathbf{x}_t) - \frac{\beta_t}{\sqrt{\alpha_t - \alpha_t\bar{\alpha}_t}}(\hat{\boldsymbol{\epsilon}}_\theta(\hat{\mathbf{x}}_t,t) - \boldsymbol{\epsilon}(\mathbf{x}_t,t))\| \\
&= \|\frac{1}{\sqrt{\alpha_t}}\Delta\mathbf{x}_t - \frac{\beta_t}{\sqrt{\alpha_t - \alpha_t\bar{\alpha}_t}}\Delta\boldsymbol{\epsilon}_t\| \\
&\leq \frac{1}{\sqrt{\alpha_t}}\|\Delta\mathbf{x}_t\| + \frac{\beta_t}{\sqrt{\alpha_t - \alpha_t\bar{\alpha}_t}}\|\Delta\boldsymbol{\epsilon}_t\|,
\end{aligned} \tag{44}$$

and rectify these two terms separately to block the propagation of errors to the next timestep. Firstly, they propose **N**oise **E**stimation **R**econstruction (NER) to linearly reconstruct the noise estimation from the quantized outputs:

$$\tilde{\boldsymbol{\epsilon}}_\theta(\hat{\mathbf{x}}_t,t) = \mathbf{K}_t \cdot \hat{\boldsymbol{\epsilon}}_\theta(\hat{\mathbf{x}}_t,t), \tag{45}$$

$$\mathrm{rQNSR}(\hat{\boldsymbol{\epsilon}}_{t,i}, \boldsymbol{\epsilon}_{t,i}) = \sqrt{\frac{\sum_j^H \sum_k^W (\hat{\boldsymbol{\epsilon}}_{t,i,j,k} - \boldsymbol{\epsilon}_{t,i,j,k})^2}{\sum_j^H \sum_k^W \boldsymbol{\epsilon}_{t,i,j,k}^2}}, \tag{46}$$

$$\begin{aligned}
\mathcal{L}(\mathbf{K}_{t,i}, \hat{\boldsymbol{\epsilon}}_{t,i}, \boldsymbol{\epsilon}_{t,i}) &= \lambda_1 \cdot \mathrm{rQNSR}(\tilde{\boldsymbol{\epsilon}}_{t,i}, \boldsymbol{\epsilon}_{t,i})^2 + \lambda_2 \cdot (\mathbf{K}_{t,i} - 1)^2, \\
&\quad + (1 - \lambda_1) \cdot \mathrm{MSE}(\tilde{\boldsymbol{\epsilon}}_{t,i}, \boldsymbol{\epsilon}_{t,i})^2,
\end{aligned} \tag{47}$$

where $t$ denotes the time step, $i$ represents the channel index, $H$ and $W$ refer to the height and width of the noise estimation, respectively. By assigning both $\lambda_1$ and $\lambda_2$ as positive coefficients, Eqn. (47) is transformed into a convex optimization problem, allowing the closed-form optimal solution for $\mathbf{K}_{t,i}$ to be precomputed. To mitigate input discrepancies, they further introduce **I**nput **B**ias **C**orrection (IBC), which caculates the average element-wise bias in the deviated model inputs over a mini-batch of $S$ samples:

$$\mathbf{B}_{t,i,j,k} = \frac{1}{S}\sum_{s=0}^{S}(\hat{\mathbf{x}}_{t,s,i,j,k} - \mathbf{x}_{t,s,i,j,k}), \quad \mathbf{B} \in \mathbb{R}^{T \times C \times H \times W}. \tag{48}$$

Finally, the corrected sampling process during inference is formalized as:

$$\tilde{\mathbf{x}}_{t-1} = \frac{1}{\sqrt{\alpha_t}}\left(\hat{\mathbf{x}}_t - \frac{\beta_t}{\sqrt{1-\bar{\alpha}_t}}\mathbf{K}_t\hat{\boldsymbol{\epsilon}}_\theta(\hat{\mathbf{x}}_t - \mathbf{B}_t, t) - \mathbf{B}_t\right) + \mathbf{z}, \tag{49}$$

where $\mathbf{z} \sim \mathcal{N}(\mathbf{0},\mathbf{I})$ represents standard Gaussian noise.

Chu et al. [107] instead decompose the quantization noise into **intra** quantization noise and **inter** quantization noise, with an analysis based on DDPM sampling as follows:

$$\begin{aligned}
\tilde{\mathbf{x}}_{t-1} &= \frac{1}{\sqrt{\alpha_t}}\left(\tilde{\mathbf{x}}_t - \frac{\beta_t}{\sqrt{1-\alpha_t}}\tilde{\boldsymbol{\epsilon}}_\theta(\tilde{\mathbf{x}}_t,t)\right) + \sigma_t\mathbf{z}, \quad \mathbf{z} \sim \mathcal{N}(0,\boldsymbol{I}) \\
&= \frac{1}{\sqrt{\alpha_t}}\left(\tilde{\mathbf{x}}_t - \frac{\beta_t}{\sqrt{1-\alpha_t}}(\boldsymbol{\epsilon}_\theta(\tilde{\mathbf{x}}_t,t) + q_\theta(\tilde{\mathbf{x}}_t,t))\right) + \sigma_t\mathbf{z}.
\end{aligned} \tag{50}$$

In the U-Net, the discrepancies in output features between the quantization modules and their full-precision counterparts are characterized as intra quantization noise. This progressively accumulates across hierarchical layers, ultimately transitioning into inter quantization noise $q_\theta(\tilde{\mathbf{x}}_t,t)$, which propagates to $\tilde{\mathbf{x}}_{t-1}$. Through empirical analysis, Chu et al. identify that during the embedding integration stage described in Eqn. (51), the embedding amplifies outliers within the stable activation distribution of normalized $h_t$, causing imbalance among channels. This ultimately manifests as the periodic amplification of intra quantization noise in a single denoising step.

$$scale_t, shift_t = layer_{emb}(emb_t).split(),$$
$$h_t = norm(h_t) * (1 + scale_t) + shift_t. \quad (51)$$

To mitigate this, they precompute static channel-wise smoothing factor $S$ to adjust the activations:

$$S = \frac{1}{T}\sum_{t=1}^{T}|1 + scale_t|, \quad (52)$$

$$Y = Q(h_t) * Q(W) = Q\left(\frac{h_t}{S}\right) * Q(SW). \quad (53)$$

Here, $\mathcal{Q}(\cdot)$ denotes the quantization operation. Furthermore, they propose a dynamic noise estimation approach to filter out inter quantization noise. Given the noisy input $\tilde{\mathbf{x}}_t$, the filtered output $\tilde{\mathbf{x}}_{t-1}$ is computed using the sampling formulation, which incorporates an optimized smoothing factor into the noise estimation network $\tilde{\epsilon}_\theta$. The noise estimation is then performed as follows:

$$\hat{\mathbf{x}}_t = \sqrt{\alpha_t}\tilde{\mathbf{x}}_{t-1} + \sqrt{1 - \alpha_t}\mathbf{z}_1, \quad (54)$$

$$\tilde{\epsilon}_\theta(\hat{\mathbf{x}}_t, t) = \epsilon_\theta(\hat{\mathbf{x}}_t, t) + q_\theta(\hat{\mathbf{x}}_t, t) \approx \mathbf{z}_1 + q_\theta(\hat{\mathbf{x}}_t, t), \quad (55)$$

$$q_\theta(\tilde{\mathbf{x}}_t, t) \approx q_\theta(\hat{\mathbf{x}}_t, t) \approx \tilde{\epsilon}_\theta(\hat{\mathbf{x}}_t, t) - \mathbf{z}_1, \quad (56)$$

where $\mathbf{z}_1$ represents standard Gaussian noise sampled randomly. By substituting Eqn. (56) into Eqn. (50), the noise-filtered $\tilde{\mathbf{x}}_{t-1}$ is obtained.

At a fine-grained level, Pandey et al. [109] identify the bias introduced by softmax quantization as a significant factor contributing to intra-quantization noise and effectively mitigate it using Absorbing Bias Correction.

### 3.1.2 QAT-Based Approaches

Quantization-Aware Training (QAT) leverages simulated quantization during training to accurately model discretization errors, enabling effective adaptation and superior performance in low-bit precision settings (e.g., 4 bits or fewer). Earlier, the seminal work LSQ established a fine-tuning paradigm that jointly optimizes quantization parameters and model weights in general domains. Subsequently, various quantization fine-tuning paradigms [81, 82, 122] have demonstrated outstanding performance in the domain of large language models (LLMs). Recently, Chang et al. [123] demonstrate an effective foundational framework for quantization fine-tuning in diffusion models, leveraging quantization distillation with a global task loss to jointly optimize quantization parameters and model weights, aligning the outputs of full-precision and quantized models. Furthermore, we highlight two significant differences between this approach and PTQ: (1) QAT typically optimizes a global task loss to guide the overall training of all learnable parameters, whereas PTQ focuses on local loss optimization; (2) QAT optimizes not only quantization parameters but also all or part of the model weights. Evidently, QAT essentially trades training overhead for enhanced model performance. Accordingly, some QAT approaches for diffusion models focus on addressing the challenges outlined in Sec. 2.3 to fully exploit the potential of learnable parameters, while others incorporate Low-Rank Adaptation (LoRA) [124] techniques to reduce the training overhead of QAT. Below, we provide a detailed overview of the latest advancements in this field.

**Holistic QAT Optimization.** Li et al. [120] first identify that the bottlenecks of low-bit quantized diffusion models (DMs) arise from large distribution oscillations in activations (**C#1**) and the accumulation of quantization errors across timesteps (**C#2**). To mitigate convergence challenges arising from distribution oscillations during training, they propose timestep-aware quantization (TaQ) and integrate it into the attention module. For any timestep $t$, TaQ collects statistics to compute the mean and variance of the query (key) activations $\mathbf{a}_q (\mathbf{a}_k)$, followed by normalization:

$$\gamma_{*;t} = \sum_{i=1}^{B} \frac{1}{b_i} \sum_{j=1}^{b_i} \mathbf{a}_*(\mathbf{x}_{t_j}, t_j), \tag{57}$$

$$\sigma_{*;t}^2 = \sum_{i=1}^{B} \frac{1}{b_i} \sum_{j=1}^{b_i} \left[\mathbf{a}_*(\mathbf{x}_{t_j}, t_j) - \gamma_{*;t}\right]^2, \tag{58}$$

$$\text{TaQ}(\mathbf{a}_*) = \lfloor \text{clip}([* - \gamma_{*;t}]/[s^* \cdot \sqrt{\sigma_{*;t}^2 + \psi}], -2^{b-1}, 2^{b-1} - 1) \rfloor. \tag{59}$$

Here, * can represent $q$ or $k$, with $B$ as the number of collected statistical data batches, $b_i$ as the batch size, and $\psi$ as a constant added to avoid 0 denominator. After smoothing, the timestep-aware quantized attention is given by:

$$\begin{aligned}
\hat{\mathbf{a}}_q(\mathbf{x}_t, t) &= s^{\mathbf{a}_q(x_t, t)} \cdot \text{TaQ}(\mathbf{a}_q(\mathbf{x}_t, t)), \\
\hat{\mathbf{a}}_k(\mathbf{x}_t, t) &= s^{\mathbf{a}_k(x_t, t)} \cdot \text{TaQ}(\mathbf{a}_k(\mathbf{x}_t, t)), \\
\mathbf{A}(\mathbf{x}_t, t) &= \text{softmax}[(\hat{\mathbf{a}}_q(\mathbf{x}_t, t) \cdot \hat{\mathbf{a}}_k(\mathbf{x}_t, t)^\top)/\sqrt{d}], \\
\hat{\mathbf{A}}(\mathbf{x}_t, t) &= s^{A(\mathbf{x}_t, t)} \cdot \text{TaQ}(\mathbf{A}(\mathbf{x}_t, t)), \\
\mathbf{a}_{\text{out}}(\mathbf{x}_t, t) &= \hat{\mathbf{A}}(\mathbf{x}_t, t) \cdot \hat{\mathbf{a}}_v(\mathbf{x}_t, t)^\top.
\end{aligned} \tag{60}$$

To alleviate error accumulation, Li et al. introduce **N**oise-**e**stimating **M**imicking (NeM), which, by incorporating the DDPM training objective, derives the final objective for Quantization-Aware Training:

$$\begin{aligned}
\arg\min_{\theta^Q} &\mathcal{L}_{\text{NeM}}(\theta^Q, \theta^{\text{FP}}) \\
&:= \mathbb{E}_{t, \mathbf{x}_0, \epsilon}[\|\epsilon_{\theta^{\text{FP}}}(\sqrt{\bar{\alpha}_t} x_0 + \sqrt{1 - \bar{\alpha}_t}\epsilon, t) - \epsilon_{\theta^Q}(\sqrt{\bar{\alpha}_t}\mathbf{x}_0 + \sqrt{1 - \bar{\alpha}_t}\epsilon, t)\|^2],
\end{aligned} \tag{61}$$

where $\theta^{\text{FP}}$ and $\theta^Q$ denote the weights of the full-precision and quantized models. Arrive here, this work validates the quantization distillation paradigm's effectiveness in low-bit diffusion model quantization, paving the way for further advancements.

Wang et al. [55] revisit the failure of PTQ under low-bit quantization and identify the root cause as the breakdown of the theoretical guarantee [47, 125] derived from Taylor expansion in Eqn. (62), where excessively large quantization perturbations $\Delta$ render it ineffective.

$$\begin{aligned}
\mathbb{E}[L(z_{n,t} + \Delta; \mathbf{w})] - \mathbb{E}[L(z_{n,t}; \mathbf{w})] &\approx \Delta^T \overline{\mathbf{g}}^{(z_{n,t})} + \frac{1}{2}\Delta^T \overline{\mathbf{H}}^{(z_{n,t})} \Delta, \\
&= \frac{1}{2}(\tilde{z}_{n,t} - z_{n,t})^T \overline{\mathbf{H}}^{(z_{n,t})} (\tilde{z}_{n,t} - z_{n,t}).
\end{aligned} \tag{62}$$

Here, $z_{n,t}$ and $\tilde{z}_{n,t}$ denote the $n$-th layer activation at timestep $t$ in the full-precision model and its quantized counterpart, respectively, with $\overline{\mathbf{g}}^{(z)}$ is the gradient and $\overline{\mathbf{H}}^{(z_{n,t})}$ is the Hessian matrix. Therefore, they linearly partition $\Delta$ into a series of small perturbations $\epsilon$, leading to:

**Theorem 2.2.**: *Given an $n$ layer diffusion model at time $t$ with quantized activations as $\tilde{\mathbf{z}}_t = [\tilde{z}_{1,t}, \tilde{z}_{2,t}, \ldots, \tilde{z}_{n,t}]$ and $\tilde{z}_{n,t} = z_{n,t} + \Delta$, where $z_{n,t}$ is the ground-truth and $\Delta$ is the large perturbation caused by low-bit quantization. Denote the target task MSE loss as $L(\tilde{z}_{n,t}, z_{n,t})$, the quantization error can be transformed into:*

$$\begin{aligned}
\mathbb{E}[L(z_{n,t} + \Delta; \mathbf{w})] - \mathbb{E}[L(z_{n,t}; \mathbf{w})] &\approx 2\epsilon^T \sum_{i=1}^{K} \left(\tilde{z}_{n-1,t}^i \cdot \mathbf{w}_n - z_{FP}\right) \\
&+ \frac{1}{2} \sum_{i=1}^{K} \left(\tilde{z}_{n,t}^i - z_{FP}\right)^T \overline{\mathbf{H}}^{(z_{n,t} + (i-1)\epsilon)} \left(\tilde{z}_{n,t}^i - z_{FP}\right),
\end{aligned} \tag{63}$$

where $\mathbf{w}_n$ is the weight for layer $n$ and $\tilde{z}_{n,t}^i = \tilde{z}_{n-1,t}^i \cdot \mathbf{w}_n$, $\epsilon$ is a perturbation small enough for Taylor expansion, $K$ is a constant, and $\Delta = K\epsilon$.

Building on this, they propose Progressive Selective Fine-Tuning (QuEST) to enhance robustness against quantization perturbations by fine-tuning model weights, effectively mitigating the long-tailed activation distribution. Specifically, they also focus on the dynamic nature of attention activation distributions (**C#1**) and the confusion of temporal information (**C#3**), leveraging temporal dynamic quantization [45] to progressively optimize temporal alignment objectives, attention alignment objectives, and the overall task objective. This work inspires the approach of prioritizing fine-tuning for quantization-sensitive modules, followed by joint tuning of all modules to enhance the model's robustness against quantization perturbations, with the main trade-off being greater procedural complexity.

Ryu et al. [56] build upon advanced techniques in LLMs [82, 81] to propose TuneQDM, a lightweight quantization fine-tuning framework. By integrating Q-Diffusion, PEQA, and DreamBooth, they establish a diffusion model fine-tuning baseline, uncovering temporal activation oscillations (**C#1**) and significant, previously unexplored shifts in inter-channel weight patterns. A key innovation lies in introducing the Multi-Channel-Wise Scale Update, which effectively adjusts channel patterns. Given the quantized weights $W_q \in \mathbb{R}^{n \times m}$ (for conv2d layers, $W_q \in \mathbb{R}^{n \times m \times k \times k}$) after initialization [43], along with intra and inter per-channel quantization scales $s_{\text{out}} \in \mathbb{R}^m$ and $s_{\text{in}} \in \mathbb{R}^n$, the fine-tuned weights $W_{\text{tuned}}$ are derived by jointly optimizing $s_{\text{in}}$ and $s_{\text{out}}$:

$$W_{\text{tuned}} = (s_{\text{out}} + \Delta s_{\text{out}}) \cdot (W_q^* - z^*) \cdot (s_{\text{in}} + \Delta s_{\text{in}}). \tag{64}$$

Here, $\Delta s_{\text{in}} \in \mathbb{R}^n$ and $\Delta s_{\text{out}} \in \mathbb{R}^m$ represent the gradient updates generated during fine-tuning for the downstream task. * indicates the frozen parameters.

**Ultra-Low-Bit DMs.** Wang et al. [59] make a groundbreaking contribution by replacing the evolving distribution in latent diffusion models from Gaussian to Bernoulli, effectively validating the remarkable compression and expressive potential of binary latent spaces. Zheng et al. [60] pioneer the exploration of weight binarization, highlighting its detrimental effect on exacerbating representation collapse and sampling divergence, which stem from the extreme amplification of the error accumulation effect (**C#2**). To mitigate these issues, they propose the **E**volvable-**B**asis **B**inarizer (EBB) and **L**ow-**R**ank **R**epresentation **M**imicking (LRM) techniques. In EBB, higher-order residual multi-basis with regularization penalties is first used to enhance the vanilla binarization:

$$\boldsymbol{w}_{\text{EBB}}^{\text{bi}} = \sigma_{\text{I}} \operatorname{sign}(\boldsymbol{w}) + \sigma_{\text{II}} \operatorname{sign}(\boldsymbol{w} - \sigma_{\text{I}} \operatorname{sign}(\boldsymbol{w})), \tag{65}$$

where the $\sigma_{\text{I}}$ and $\sigma_{\text{II}}$ are learnable scalars, initialized as $\sigma_{\text{I}}^0 = \frac{\|\boldsymbol{w}\|}{n}$ and $\sigma_{\text{II}}^0 = \frac{\|\boldsymbol{w} - \sigma_{\text{I}} \operatorname{sign}(\boldsymbol{w})\|}{n}$, respectively, $\|\cdot\|$ denotes the $\ell_2$-normalization. then transitions into the second stage with simple single-basis binary weights. Subsequently, the second term is discarded, directly optimizing the simple single-basis binary weights. In LRM, considering the limited representational capacity of the binary space, intermediate layer features are first decomposed via PCA, followed by the fine-grained alignment of full-precision components and their quantized counterparts in the low-rank space. Ultimately, BinaryDM advances quantization precision to W1A4.

Zheng et al. [61] propose BiDM, achieving the theoretical limit of quantization with W1A1. To address temporal dynamic activations (**C#1**), they introduce the **T**imestep-friendly **B**inary **S**tructure (TBS), which incorporates two key enhancements to the prior design. The first involves making the matrix $K$ in the original XNOR [126] convolution learnable:

$$\begin{aligned}\mathbf{I} * \mathbf{W} &\approx (\operatorname{sign}(\mathbf{I}) \otimes \operatorname{sign}(\mathbf{W})) \odot (K\alpha) \\ &= (\operatorname{sign}(\mathbf{I}) \otimes \operatorname{sign}(\mathbf{W})) \odot (A * k\alpha),\end{aligned} \tag{66}$$

$$\frac{\partial \mathcal{L}}{\partial k} = \frac{\partial \mathcal{L}}{\partial (\mathbf{I} * \mathbf{W})} \cdot \frac{\partial (A * k\alpha)}{\partial k} (\operatorname{sign}(\mathbf{I}) \otimes \operatorname{sign}(\mathbf{W})), \tag{67}$$

where $\mathbf{I} \in \mathbb{R}^{c \times w_{\text{in}} \times h_{\text{in}}}$, and $\mathbf{W} \in \mathbb{R}^{c \times w \times h}$, $A = \frac{\sum |\mathbf{I}_{i,:,:}|}{c}$, $\alpha = \frac{1}{n}\|\mathbf{W}\|_{\ell_1}$ and $k \in \mathbb{R}^{1 \times 1 \times w \times h}$ represents a 2D filter; furthermore, $*$ and $\otimes$ indicate convolution with and without multiplication, respectively. Secondly, it incorporates learnable momentum updates within the DeepCache [127] mechanism to further enhance compensatory representation capabilities:

$$\operatorname{Concat}\left(D_m^{t-1}(\cdot), (1 - \alpha_{m+1}^{t-1}) \cdot U_{m+1}^{t-1}(\cdot) + \alpha_{m+1}^{t-1} \cdot U_{m+1}^t(\cdot)\right), \tag{68}$$

where $\alpha_{m+1}^{t-1}$ is a learnable scaling factor. Then, to overcome the challenge of binary-to-full-precision alignment, they apply **S**pace **P**atched **D**istillation (SPD) to the intermediate features.

Specifically, the intermediate features $\mathcal{F}^{\text{fp}}$ and $\mathcal{F}^{\text{bi}} \in \mathbb{F}^{b \times c \times w \times h}$, output by the full-precision and quantized blocks respectively, are first divided into $p^2$ patches:

$$\mathcal{P}_{i,j}^{\text{fp}} = \mathcal{F}_{[:,:,i:i+w/p,j:j+h/p]}^{\text{fp}}, \quad \mathcal{P}_{i,j}^{\text{bi}} = \mathcal{F}_{[:,:,i:i+w/p,j:j+h/p]}^{\text{bi}}, \quad (69)$$

Then, attention-guided loss is computed for each patch separately, followed by regularization and accumulation to obtain the distillation loss $\mathcal{L}_{\text{SPD}}^m$, ultimately leading to the final loss $\mathcal{L}$:

$$\mathcal{A}_{i,j}^{\text{fp}} = \mathcal{P}_{i,j}^{\text{fp}} \mathcal{P}_{i,j}^{\text{fp}\,T}, \quad \mathcal{A}_{i,j}^{\text{bi}} = \mathcal{P}_{i,j}^{\text{bi}} \mathcal{P}_{i,j}^{\text{bi}\,T}, \quad (70)$$

$$\mathcal{L}_{\text{SPD}}^m = \frac{1}{p^2} \sum_{i=0}^{p-1} \sum_{j=0}^{p-1} \left\| \frac{\mathcal{A}_{i,j}^{\text{fp}}}{\|\mathcal{A}_{i,j}^{\text{fp}}\|_2} - \frac{\mathcal{A}_{i,j}^{\text{bi}}}{\|\mathcal{A}_{i,j}^{\text{bi}}\|_2} \right\|_2, \quad (71)$$

$$\mathcal{L} = \mathcal{L}_{\text{DM}} + \frac{\lambda}{2d+1} \sum_{m}^{2d+1} \mathcal{L}_{\text{SPD}}^m, \quad (72)$$

where $d$ denotes the number of blocks in the upsampling process or downsampling process, and $\lambda$ is a regularization coefficient with a default value of 4.

Sui et al. [62] introduce BitsFusion, a comprehensive 1.99-bit weight quantization framework tailored for large-scale text-to-image diffusion models. It incorporates adaptive mixed-precision optimization, optimal initialization for low-bit DMs, and an enhanced two-stage training pipeline, offering a highly effective solution for extreme low-bit quantization. Chen et al. [63] introduce BI-DiffSR, a novel binarized diffusion model for image **S**uper-**R**esolution (SR). It first overcomes challenge **C#4** by optimizing the well-binarized Unet structure, and then improves the expressive power of the binary module through **T**imestep-**a**ware **R**edistribution (TaR) and **A**ctivation function (TaA), effectively alleviating challenge **C#1**.

**LoRA-Based Enhancements.** LoRA [124, 128, 129] achieves efficient fine-tuning by constraining the model parameter updates to possess a low intrinsic rank, denoted as $r$. Given a pretrained linear module $\mathbf{Y} = \mathbf{X}\mathbf{W}_0$, where $\mathbf{X} \in \mathbb{R}^{b \times c_{in}}$ and $\mathbf{W}_0 \in \mathbb{R}^{c_{in} \times c_{out}}$, with $b$ representing the batch size, $c_{in}$ and $c_{out}$ representing the number of input and output channels, respectively, LoRA fixes the original weights $\mathbf{W}_0$ and introduces updates as follows:

$$\mathbf{Y} = \mathbf{X}\mathbf{W}_0 + \mathbf{X}\mathbf{B}\mathbf{A}, \quad (73)$$

where $\mathbf{B} \in \mathbb{R}^{c_{in} \times r}$ and $\mathbf{A} \in \mathbb{R}^{r \times c_{out}}$ are the learnable two low-rank matrices with $r \ll \min(c_{in}, c_{out})$.

Nevertheless, this approach incurs limitations when both weights and activations are quantized, denoted by $\hat{\mathbf{W}}_0$ and $\hat{\mathbf{X}}$, respectively. Inspired by QLoRA [130], He et al. [57] introduce a novel LoRA-based quantization-aware and parameter-efficient fine-tuning framework for low-bit diffusion models, dubbed EfficientDM, to achieve QAT-level performance with PTQ-like efficiency. To circumvent the direct matrix multiplication between $\hat{\mathbf{X}}$ and the full-precision matrix $\mathbf{B}\mathbf{A}$, they first merges $\mathbf{B}\mathbf{A}$ with $\mathbf{W}_0$, followed by joint quantization:

$$\mathbf{Y} = \mathcal{Q}_U(\mathbf{X}, s_x) \mathcal{Q}_U(\mathbf{W}_0 + \mathbf{B}\mathbf{A}, s_w) = \hat{\mathbf{X}}\hat{\mathbf{W}}, \quad (74)$$

where $s_w$ denotes the channel-wise quantization scale for weights and $s_x$ is the layer-wise quantization scale for activations. Subsequently, the parameters $s_w, s_x$ and $\mathbf{B}\mathbf{A}$ are simultaneously optimized using the MSE loss:

$$\mathcal{L}_t = \|\boldsymbol{\epsilon}_\theta(\mathbf{x}_t, t) - \hat{\boldsymbol{\epsilon}}_\theta(\mathbf{x}_t, t)\|^2. \quad (75)$$

To further enhance the optimization of the LoRA weights, the ratio $R = \frac{\nabla_{\mathbf{BA}}\mathcal{L}}{\overline{s_w}}$ is maintained approximately constant across each layer, where $\overline{s_w}$ represents the averaged weight quantization scale across channels.

Zeng et al. [131] further analyze the interference effects of quantization errors on sampling acceleration, in conjunction with sampling acceleration theory. They propose aligning high-order sampling trajectories of full precision with low-order quantized sampling trajectories through **M**ixed-**O**rder **T**rajectory **A**lignment (MOTA), enabling the learning of a more linear probabilistic flow to better tolerate quantization disturbances. Building on this, they apply the sampling-aware approach to QLoRA, resulting in the **S**ampling-**A**ware QLoRA (SA-QLoRA).

Guo et al. propose IntLoRA [58], a parameter-efficient fine-tuning (PEFT) approach for large-scale text-to-image models. Unlike conventional fine-tuning methods that require PTQ to merge quantized pretrained model weights $\hat{\mathbf{W}}$ with full-precision LoRA weights $\mathbf{AB}$, IntLoRA directly trains an integer-only $\hat{\mathbf{AB}}$, enabling PTQ-free fusion with pretrained weights that are independently quantized. It first applies **A**daptation-**Q**uantization **S**eparation (AQS), which enables a fine-tuning-friendly initialization of $\hat{\mathbf{AB}}$ with all-zero values, formulated as:

$$\mathbf{W}' = \mathcal{Q}[\mathbf{W} - \text{sg}(\mathbf{R})] + \text{sg}(\mathbf{R}) + \hat{\mathbf{AB}}, \tag{76}$$

where $\mathbf{R}$ is the non-zero auxiliary matrix, and $\text{sg}(\cdot)$ denotes the stop gradient operation. Then, **M**ultiplicative **L**ow-Rank **A**daptation (MLA) is used to achieve seamless fusion of $\hat{\mathbf{W}}$ and $\hat{\mathbf{AB}}$. Specifically, denote the quant-dequant results as $\mathcal{Q}(\mathbf{W} - \mathbf{R}) = s \cdot (\mathbf{W}_{round} - z)$, then the MLA can be derived as follows:

$$\begin{aligned}
\mathbf{W}' &= \mathcal{Q}(\mathbf{W} - \mathbf{R}) + \mathbf{R} + \hat{\mathbf{AB}}, \\
&= s \cdot (\mathbf{W}_{round} - z) + \mathbf{R} + \hat{\mathbf{AB}}, \\
&= [s \cdot \mathbf{I} + \frac{1}{\mathbf{W}_{round} - z} \odot (\mathbf{R} + \hat{\mathbf{AB}})] \odot (\mathbf{W}_{round} - z),
\end{aligned} \tag{77}$$

where the operator $\odot$ denotes the Hadamard product of two matrices. Finally, the **V**ariance **M**atching **C**ontrol (VMC) mechanism is introduced to adjust the distribution of $\mathbf{R}$.

### 3.2 Diffusion Transformer Quantization

#### 3.2.1 PTQ-Based Approaches

Different from conventional U-Net-based diffusion models, Peebles et al. [65] recently introduce a novel variant, the Diffusion Transformer (DiT), which employs a transformer-only architecture to predict $\boldsymbol{\epsilon}_\theta(\mathbf{x}_t, t)$ and $\boldsymbol{\Sigma}_\theta(\mathbf{x}_t, t)$ for sampling. Building upon the transformer's prowess in capturing long-range dependencies and its scaling capabilities, DiT has shown remarkable potential in image generation. Nevertheless, quantizing DiT is challenging due to both variable data variance within the transformer blocks [132–134, 49] (**C #6**) and distribution shifts across timesteps (**C #1**). In the following, we categorize and review current DiT quantization efforts based on their core techniques.

**Group-wise Quantization.** The granularity of quantization determines the scale of data sharing quantization parameters. Coarse-grained tensor-wise quantization incurs lower computational overhead but is susceptible to local outliers, whereas fine-grained channel-wise quantization can isolate outliers from other channels at the cost of higher computational complexity. To balance precision and computational efficiency, a compromise has been proposed: group-wise quantization [135], which aggregates multiple channels into a group for parameter sharing. Yang et al. [66] first analyze activation patterns and error accumulation under various calibration strategies, proposing 1-step calibration—using samples from the first sampling step—to address **C #1**. In analyzing weight patterns, they identify high discreteness and numerous outliers within individual channels, prompting the proposal of Weight Group Quantization to mitigate outlier impact by aggregating multiple channels. Further, Chen et al. [67] observe that the variance across input channels of weight is much more significant than the output channels. They argued that input channels, which handle correlations between different features of all input tokens, are more critical than output channels, which process the same feature across different tokens, and therefore propose quantizing the weights along with the input channels. Additionally, to mitigate the impact of structured outliers within channels, they employ group quantization. Noting that group size exhibits non-monotonic behavior, they further introduce Automatic Group Allocation, which determines the optimal group size $\mathbf{g}^*$ by minimizing the FID score between the true sample $\mathbf{R}$ and the samples generated by the quantized model $\mathbf{G}$. This process is formalized as:

$$L(\hat{\mathbf{w}}, \mathbf{g}) = \text{FID}(\mathbf{R}, \mathbf{G}_{\hat{\mathbf{w}}, \mathbf{g}}), \tag{78}$$

$$\mathbf{g}^* = \arg\min_{\mathbf{g}} L(\hat{\mathbf{w}}, \mathbf{g}), \quad \text{s.t. } B(\mathbf{g}) \le N_{bitops}, \tag{79}$$

where $\hat{\mathbf{w}}$ is the quantized weight, $B(\cdot)$ is the measurement of bit-operations (BitOps), and $N_{bitops}$ is the predefined threshold.

**Channel Equalization.** Channel equalization [136–138, 74] is a reparameterization technique that enhances the model's robustness to quantization. Ideally, the ranges of each channel are equal to the total range of the weight tensor, meaning we use the best possible representative power per channel. If unmet, the range can be redistributed between the activation $\mathbf{X}$ and weight $\mathbf{W}$ by adjusting the scaling factors. For instance, the linear transformation $\mathbf{Y} = \mathbf{X}\mathbf{W}$ can be reparameterized as $\mathbf{Y} = \mathbf{X}\mathbf{S}_\mathbf{x} \cdot \mathbf{S}_\mathbf{w}\mathbf{W}$, where $\mathbf{S}_\mathbf{x}$ and $\mathbf{S}_\mathbf{w}$ are scaling factors satisfying $\mathbf{S}_\mathbf{x} \cdot \mathbf{S}_\mathbf{w} = \mathbf{I}$ to ensure numerical consistency. Wu et al. [68] first identify significant outliers in the weight and activation channels of DiT's linear layers. To mitigate substantial quantization errors from direct clipping, they propose **C**hannel-wise **S**alience **B**alancing (CSB), utilizing diagonal matrices $\mathbf{B}^\mathbf{X}$ and $\mathbf{B}^\mathbf{W}$ to balance channel-wise distributions of a token sequence $\mathbf{X} \in \mathbb{R}^{n \times d_{in}}$ and weight matrix $\mathbf{W} \in \mathbb{R}^{d_{in} \times d_{out}}$, expressed as:

$$\widetilde{\mathbf{X}} = \mathbf{X}\mathbf{B}^\mathbf{X}, \ \widetilde{\mathbf{W}} = \mathbf{W}\mathbf{B}^\mathbf{W}. \tag{80}$$

To achieve balanced distributions in $\mathbf{X}$ and $\mathbf{W}$, $\mathbf{B}^\mathbf{X}$ and $\mathbf{B}^\mathbf{W}$ must effectively capture the characteristics of salient channels. The *salience s* of a channel is measured as the maximum absolute value among its elements, based on which $\mathbf{B}^\mathbf{X}$ and $\mathbf{B}^\mathbf{W}$ are computed as follows:

$$s(\mathbf{X}_j) = \max(|\mathbf{X}_j|), \ s(\mathbf{W}_j) = \max(|\mathbf{W}_j|), \quad \text{where} \quad j = 1, 2, \ldots, d_{in}, \tag{81}$$

$$\widetilde{s}(\mathbf{X}_j, \mathbf{W}_j) = (s(\mathbf{X}_j) \cdot s(\mathbf{W}_j))^{\frac{1}{2}}, \tag{82}$$

$$\mathbf{B}^\mathbf{X} = \mathrm{diag}\left(\frac{\widetilde{s}(\mathbf{X}_1, \mathbf{W}_1)}{s(\mathbf{X}_1)}, \frac{\widetilde{s}(\mathbf{X}_2, \mathbf{W}_2)}{s(\mathbf{X}_2)}, \ldots, \frac{\widetilde{s}(\mathbf{X}_{d_{in}}, \mathbf{W}_{d_{in}})}{s(\mathbf{X}_{d_{in}})}\right), \tag{83}$$

$$\mathbf{B}^\mathbf{W} = \mathrm{diag}\left(\frac{\widetilde{s}(\mathbf{X}_1, W_1)}{s(\mathbf{W}_1)}, \frac{\widetilde{s}(\mathbf{X}_2, \mathbf{W}_2)}{s(\mathbf{W}_2)}, \ldots, \frac{\widetilde{s}(\mathbf{X}_{d_{in}}, \mathbf{W}_{d_{in}})}{s(\mathbf{W}_{d_{in}})}\right), \tag{84}$$

Second, to accurately gauge the activation channel salience under multi-timestep scenarios, Wu et al. propose **S**pearmen's $\rho$-guided **S**alience **C**alibration (SSC). Given an activation sequence $\mathbf{X}^{(1:T)} = (\mathbf{X}^{(1)}, \mathbf{X}^{(2)}, \ldots, \mathbf{X}^{(T)})$ spanning $T$ timesteps, the salience of all activation and weight channels at a specific timestep $t$ can be estimated using Eqn. (81):

$$\mathbf{s}(\mathbf{X}^{(t)}) = (s(\mathbf{X}_1^{(t)}), s(\mathbf{X}_2^{(t)}), \ldots, s(\mathbf{X}_{d_{in}}^{(t)})), \quad \mathbf{s}(\mathbf{W}) = (s(\mathbf{W}_1), s(\mathbf{W}_2), \ldots, s(\mathbf{W}_{d_{in}})). \tag{85}$$

Building on the core insight that a lower correlation between activation salience $\mathbf{s}(\mathbf{X}^{(t)})$ and weight salience $\mathbf{s}(\mathbf{W})$ leads to a greater reduction effect in overall channel salience, they formulate the *Spearman's $\rho$-calibrated Temporal Salience $s_\rho$* by selectively aggregating the activation salience across timesteps:

$$\mathbf{s}_\rho(\mathbf{X}^{(1:T)}) = (\eta_1, \eta_2, \ldots, \eta_T) \cdot (\mathbf{s}(\mathbf{X}^{(1)}), \mathbf{s}(\mathbf{X}^{(2)}), \ldots, \mathbf{s}(\mathbf{X}^{(T)}))^\mathrm{T} \in \mathbb{R}^{d_{in}}, \tag{86}$$

where weighting factors $\{\eta_t\}_{t=1}^T$ are derived from a normalized exponential form of inverse Spearman's $\rho$ statistic [139]:

$$\eta_t = \frac{\exp[-\rho(\mathbf{s}(\mathbf{X}^{(t)}), \mathbf{s}(\mathbf{W}))]}{\sum_{\tau=1}^T \exp[-\rho(\mathbf{s}(\mathbf{X}^{(\tau)}), \mathbf{s}(\mathbf{W}))]} \tag{87}$$

Subsequently, $\mathbf{s}_\rho$ is substituted into Eqn. (80) (81) (82) to obtain $\mathbf{B}_\rho^\mathbf{X}$ and $\mathbf{B}_\rho^\mathbf{W}$, ultimately reparameterizing the linear layer as follows:

$$\widetilde{\mathbf{X}} \cdot \widetilde{\mathbf{W}} = (\mathbf{X}\mathbf{B}_\rho^\mathbf{X}) \cdot (\mathbf{B}_\rho^\mathbf{W}\mathbf{W}) = \mathbf{X} \cdot \mathbf{W}. \tag{88}$$

Similarly, Dong et al. [69] propose **T**emporal-**a**ggregated **S**moothing (TAS), which achieves channel rebalancing through a channel-wise smoothing factor that aggregates outlier magnitude information across multiple timesteps. Specifically, for a linear layer, given the input $\mathbf{X} \in \mathbb{R}^{B \times T \times L \times C}$ and weight $\mathbf{W}^{N \times C}$, where $B, T, L, C, N$ represent the batch size of the calibration data, total time steps, token length, input channels and output channels respectively, the scaling factor for the $c$-th input channel is computed as follows:

$$\mathbf{s}_c = \frac{\max_{1 \leq t \leq T, 1 \leq b \leq B, 1 \leq l \leq L}(|\mathbf{X}_{btlc}|)^\alpha}{\max_{1 \leq n \leq N}(|\mathbf{W}_{nc}|)^{1-\alpha}}. \tag{89}$$

where the hyperparameter $\alpha$ determines the extent to which we aim to shift the impact of outliers from activations to weights. Let $\mathbf{s} = [\mathbf{s}_1, \mathbf{s}_2, \ldots, \mathbf{s}_C]$, then the linear output $\mathbf{Y}_t$ is expressed as follows:

$$\mathbf{Y}_t = Q(\mathbf{X}_t \, \mathrm{diag}(\mathbf{s})^{-1})(\mathrm{diag}(\mathbf{s})\mathbf{W}). \tag{90}$$

Furthermore, to identify the most effective TAS factor, they parameterize **s** as learnable and employ Grid Search Optimization to determine the optimal $\alpha$, as shown below:

$$\mathcal{L}(\mathbf{s}) = \sum_{t=1}^{T} \|Q(\mathbf{X}_t \operatorname{diag}(\mathbf{s})^{-1})Q(\operatorname{diag}(\mathbf{s})\mathbf{W} - \mathbf{X}_t\mathbf{W})\|^2, \tag{91}$$

$$\alpha^* = \arg\min_{\alpha} \mathcal{L}\left(\frac{a_1^{\alpha}}{b_1^{(1-\alpha)}}; \frac{a_2^{\alpha}}{b_2^{(2-\alpha)}}; \cdots; \frac{a_C^{\alpha}}{b_C^{(1-\alpha)}}\right), \tag{92}$$

where $a_c = \max_{1 \leq t \leq T, 1 \leq b \leq B, 1 \leq l \leq L}(|\mathbf{X}_{btlc}|)$ and $b_c = \max_{1 \leq n \leq N}(|\mathbf{X}_{nc}|)$.

Zhao et al.[70] pioneered the first successful quantization implementation for more challenging text-to-image and video generation tasks in large-scaled DiT. They identified the primary challenges as stemming from multi-level data variation inherent in large-scaled DiT, specifically encompassing token-level variation, condition-level variation, timestep-level variation, and channel-level variation. To address these challenges, the authors proposed **V**ideo & **I**mage **D**iffusion **T**ransformer **Q**uantization (ViDiT-Q), an innovative quantization framework that employs token-wise quantization for activations and channel-wise quantization for weight parameters.

To address temporal and conditional data variation, they innovatively propose an online dynamic quantization parameter computation mechanism, which effectively mitigates error accumulation caused by conventional static quantization parameter determination methods. To address data variation within temporal groups through denoising timesteps, they identified that channel-wise activation imbalance under temporal evolution stems from feature modulation mechanisms. Specifically, the pretrained *scale-shift table* predominantly governs activation distributions during initial denoising stages, whereas time embeddings dominate activation patterns in later stages. To resolve excessive intra-channel data variation in DiT, two dedicated solutions were developed: 1) A scale-based method[74] for initial denoising stages, and 2) A rotation-based method[140, 48] for latter stages.

The scale-based approach introduces a learnable balancing mask $s$ that strategically redistributes quantization complexity from activations to weights, formulated as:

$$Y = \left(X \operatorname{diag}(s)^{-1}\right) \cdot (\operatorname{diag}(s)W) = \hat{X} \cdot \hat{W}; \qquad s_i = \frac{\max(|X_i|)^{\alpha}}{\max(|W_i|)^{1-\alpha}}, \tag{93}$$

Here, $X$, $Y$, and $W$ denote the input, output activation, and weights, respectively. $s$ represents the channel balancing mask, where $\alpha$ serves as a hyperparameter controlling the quantization equilibrium. The selection of $\alpha$ exhibits non-trivial sensitivity to quantization performance. By strategically constraining the optimization of $\alpha$ to only a few critical timesteps during the initial denoising stage, this approach effectively circumvents the prohibitive computational overhead that would otherwise arise from searching for the optimal $\alpha$ across all timesteps.

Subsequently, the rotation-based method introduces an orthogonal rotation matrix $Q$, which satisfies $QQ^T = I$ and $|Q| = 1$ Multiplying the data matrices by $Q$ on both sides preserves computational invariance, $Y = XW^T = (XQ)(Q^TW)$. This rotation operation ensures more uniform channel-wise data distribution. Since earlier application of the scale-based method has mitigated extreme channel imbalance, the rotation approach maintains balanced activation distributions for subsequent timesteps in denoising inference.

Finally, they analyzed quantization sensitivity across different DiT modules, including cross-attention layers, self-attention layers, feedforward network (FFN) layers, and temporal attention layers. Using a controlled variable approach, they correlated layer-wise sensitivity levels with changes in evaluation metrics before/after quantization. This enabled a fine-grained mixed-precision scheme: high-sensitivity layers receive high-bit precision, while low-sensitivity layers use low-bit precision.

Similarly, Liu et al.[71] proposed Hybrid Floating-point Quantization for DiT (HQ-DiT), an efficient post-training quantization framework that employs 4-bit floating-point (FP) precision for both weights and activations in DiT-based inference. This work pioneers the implementation of 4-bit FP quantization in DiTs, addressing the limited dynamic range and inflexibility of conventional fixed-point quantization schemes (e.g., INT8) while minimizing quantization error through precision-adaptive bit allocation.

Simultaneously, they apply the Hadamard transform to all layers in DiT, significantly reducing data outliers. Next, to determine the flexible trade-off between exponent bits and mantissa bits under FP4 precision, they calculate the ratio $s_w$ between the maximum value and the $\alpha$ percentile of weight channel data, avoiding interference from extreme values. They also compute the ratio $r$ of maximum-to-minimum representable values for different exponent-mantissa bit combinations in FP4. By closely approximating they dynamically determine the quantization parameters (mantissa/exponent bit allocation) under FP4, as shown below:

$$s_w = \frac{\max(|W|)}{\text{Quantile}(|W|, \alpha)}; \qquad r = 2 \times \frac{max_val}{min_val} = 2^{(2^{n_e})} \times \frac{(2 - 2^{-n_m})}{(1 + 2^{-n_m})}, \tag{94}$$

where $\alpha$ is a hyperparameter that return the bottom $\alpha$ percentile of the weight values and they set $\alpha = 25$. The $max\_val = 2^{(2^{n_e-1})} \times (2 - 2^{-n_m})$ and $min\_val = (1 + 2^{-n_m})$. For activation quantization, the authors directly employ the Min-max[43] quantization method.

Li et al. [141] propose SVDQuant, a novel 4-bit quantization paradigm. Unlike smoothing, which redistributes outliers between weights and activations, SVDQuant absorbs these outliers using a low-rank branch. The outliers are first consolidated by shifting them from activations to weights. Then, a high-precision, low-rank branch processes the weight outliers via Singular Value Decomposition (SVD), while a low-bit quantized branch handles the residuals. This method alleviates quantization challenges on both ends. However, running the low-rank branch independently incurs significant overhead due to extra data movement, which offsets the quantization speedup. To mitigate this, they co-design the Nunchaku inference engine, which integrates the kernels of the low-rank branch with those of the low-bit branch, eliminating redundant memory accesses.

### 3.2.2 QAT-Based Approaches

Current research on Diffusion Transformers reveals limited application of Quantization-Aware Training, primarily due to the substantial parameter sizes in these models, which impose significant computational and resource constraints. Nevertheless, optimizing QAT efficiency through advanced memory compression techniques presents a promising direction for future exploration. By enhancing memory utilization, it may be possible to alleviate the resource burden of QAT, thereby facilitating its more practical and efficient integration into Diffusion Transformers.

## 4 Benchmarking Experiments

### 4.1 Experimental Setup

We comprehensively evaluate the open-source PTQ-based and QAT-based methods in three tasks: unconditional image generation, class-conditional image generation, and text-conditional guided image generation. The details of the benchmarks for each task are outlined as follows.

**Quantization Settings.** Notationally, WxAy indicates that weights and activations are quantized to x and y bits, respectively. For PTQ-based methods, we conduct evaluation experiments under the W8A8 and W4A8 settings. For QAT-based methods, we assess performance under the W8A8, W4A8, and W4A4 configurations. As for calibration, if a custom strategy is available, it is applied; otherwise, we sample the calibration dataset at regular timestep intervals from the denoising pipeline.

**Class-conditional Image Generation.** We evaluate the class-conditional image generation performance of various methods on the ImageNet 256×256 [87] dataset using LDM-4 [25]. The DDIM sampler is employed with the following parameter settings: classifier-free guidance $scale = 3.0$, sampling $steps = 20$, and variance schedule $\eta = 0.0$. In each evaluation round, 50,000 images are generated, and quantitative evaluation metrics, including Inception Score (IS), Fréchet Inception Distance (FID), scaled Fréchet Inception Distance (sFID), precision, and recall, are computed using OpenAI's evaluator [142].

**Unconditional Image Generation.** We evaluate the unconditional image generation performance of various methods using LDM-4 [25] and LDM-8 [25] on the LSUN-Bedrooms 256×256 [88] and LSUN-Churches 256×256 [88] datasets. The DDIM sampler is employed with the following parameter settings: classifier-free guidance $scale = 3.0$, sampling $steps = 20$, and variance schedule $\eta = 0.0$. In each evaluation round, 50,000 images are generated, and the FID, sFID, precision, and recall are computed as quantitative evaluation metrics using OpenAI's evaluator [142].

Table 2: Performance evaluation of various approaches for class-conditional image generation, conducted using LDM-4 ($scale = 3.0, \eta = 0.0, steps = 20$) on the ImageNet 256×256 dataset. The dagger (†) symbol indicates QAT-based methods.

| Method | Bits (W/A) | Size (MB) | BOPs (T) | IS ↑ | FID ↓ | sFID ↓ | Precision ↑ | Recall ↑ |
|---|---|---|---|---|---|---|---|---|
| FP | 32/32 | 1742.72 | 102.20 | 366.03 | 11.13 | 7.834 | 93.93% | 27.98% |
| PTQ4DM [40] | 8/8 | 436.79 | 8.76 | 322.16 | 9.37 | 9.87 | 87.15% | 31.77% |
| Q-Diffusion [41] | 8/8 | 436.79 | 8.76 | 327.16 | 8.72 | 10.46 | 86.91% | 33.26% |
| EDA-DM [50] | 8/8 | 436.79 | 8.76 | 337.95 | 11.30 | 8.93 | 85.72% | 30.18% |
| TFMQ-DM [52] | 8/8 | 436.79 | 8.76 | 363.12 | 10.82 | 7.99 | **93.70%** | 29.20% |
| QNCD [107] | 8/8 | 436.79 | 8.76 | 318.23 | 11.41 | 10.75 | 82.86% | 27.92% |
| PTQD [44] | 8/8 | 436.79 | 8.76 | 324.64 | 8.46 | 10.12 | 87.68% | 34.64% |
| D$^2$-DPM [46] | 8/8 | 436.79 | 8.76 | 333.89 | **8.12** | **7.92** | 88.56% | 36.69% |
| EfficientDM † [57] | 8/8 | 436.79 | 8.76 | 316.58 | 11.62 | 10.19 | 85.24% | 36.20% |
| QuEST † [55] | 8/8 | 436.79 | 8.76 | **365.70** | 10.93 | 7.95 | 90.51% | **38.33%** |
| PTQ4DM [40] | 4/8 | 219.12 | 4.38 | 336.28 | 10.45 | 13.94 | 90.61% | 28.63% |
| Q-Diffusion [41] | 4/8 | 219.12 | 4.38 | 347.52 | 11.13 | 9.07 | 90.89% | 29.39% |
| EDA-DM [50] | 4/8 | 219.12 | 4.38 | 348.90 | 10.39 | 7.86 | 91.22% | 29.80% |
| TFMQ-DM [52] | 4/8 | 219.12 | 4.38 | 194.43 | 17.63 | 43.10 | 69.19% | 26.59% |
| QNCD [107] | 4/8 | 219.12 | 4.38 | 329.71 | 12.05 | 11.97 | 84.70% | 24.41% |
| PTQD [44] | 4/8 | 219.12 | 4.38 | 355.10 | 10.41 | 8.45 | 92.13% | 27.54% |
| D$^2$-DPM [46] | 4/8 | 219.12 | 4.38 | **358.14** | 9.75 | **6.60** | 92.25% | 30.21% |
| EfficientDM † [57] | 4/8 | 219.12 | 4.38 | 338.05 | 8.87 | 7.92 | 92.09% | 31.24% |
| QuEST † [55] | 4/8 | 219.12 | 4.38 | 342.64 | **8.52** | 6.74 | **93.28%** | **32.90%** |
| EfficientDM † [57] | 4/4 | 219.12 | 2.19 | 170.21 | 7.99 | 10.02 | 79.91% | **41.68%** |
| QuEST † [55] | 4/4 | 219.12 | 2.19 | **197.02** | **7.76** | 9.89 | **83.41%** | 34.94% |

**Text-conditional Guided Image Generation.** We evaluate the performance of text-guided image generation using Stable Diffusion v1-4 on the MS-COCO 512x512 [91] dataset. The PLMS [29] sampler is employed with 50 sampling steps, and the classifier-free guidance (cfg) is fixed at the default value of 7.5 in Stable Diffusion, serving as a trade-off between sample quality and diversity. For each evaluation round, we generate 30,000 images and compute the FID, sFID, and CLIP score [143] as quantitative evaluation metrics.

### 4.2 Benchmark Results

#### 4.2.1 Class-conditional Image Generation

Table. 2 presents the performance evaluation results of nine representative open-source solutions for class-conditional image generation. The dagger (†) symbol indicates QAT-based methods, while the remaining are PTQ-based approaches. Specifically, PTQ4DM [40], Q-Diffusion [41], and EDA-DM [50] focus on custom calibration schemes and PTQ training objectives, TFMQ-DM [52] primarily emphasizes temporal feature alignment, while QNCD [107], PTQD [44], and D$^2$-DPM [46] are dedicated to custom quantization error correction strategies. In contrast, EfficientDM [57] and QuEST [55] achieve high-fidelity quantization through fine-tuning.

**Under high-bit quantization, the PTQ approach demonstrates competitive performance.** Notably, in the W8A8 and W4A8 quantization settings, the PTQ-based approach is not substantially inferior to the QAT-based method, and in certain cases, it even outperforms it. For example, in the W8A8 quantization setting, PTQ4DM [40], PTQD [44], and D2-DPM [46] surpass the LoRA-based EfficientDM in terms of FID/sFID, with scores of 2.25/0.32, 3.16/0.07, and 3.50/2.27, respectively. This outcome not only underscores the effectiveness of the customized calibration set sampling strategy and quantization error correction techniques but also suggests that when quantization noise is minimal, extensive weight fine-tuning may not necessarily yield optimal results. In fact, excessive fine-tuning may have an adverse impact, indicating that fine-tuning should be applied judiciously to avoid diminishing returns.

**Reducing the bitwidth of activations leads to more pronounced performance degradation.** The data indicates that when the quantization setting is lowered from W8A8 to W4A8, PTQ-based methods such as Q-Diffusion and EDA-DM do not exhibit significant performance degradation. However, when the quantization setting is further reduced to W4A4, the PTQ algorithms no longer remain effective.

**The PTQ method fails to converge under 4-bit activation quantization.** A theoretical explanation for this failure is provided here. In most PTQ-based quantization schemes, the baseline BRECQ [47] is employed, which performs high-fidelity quantization through block-wise module reconstruction. This process essentially relies on the Taylor expansion theorem to approximate the loss degradation:

$$\mathbb{E}[L(\mathbf{w} + \Delta\mathbf{w})] - \mathbb{E}[L(\mathbf{w})] \approx \Delta\mathbf{w}^\mathrm{T}\bar{\mathbf{g}}^{(\mathbf{w})} + \frac{1}{2}\Delta\mathbf{w}^\mathrm{T}\bar{\mathbf{H}}^{(\mathbf{w})}\Delta\mathbf{w}, \tag{95}$$

where $\bar{\mathbf{g}}^{(\mathbf{w})} = \mathbb{E}[\nabla_\mathbf{w} L]$ and $\bar{\mathbf{H}}^{(\mathbf{w})} = \mathbb{E}[\nabla_\mathbf{w}^2 L]$ are the gradients and the Hessian matrix and $\Delta\mathbf{w}$ is the weight perturbation. Denote the neural network output $\mathbf{z}^{(n)} = f(\theta) \in \mathcal{R}^m$, the loss function can be represented by $L(f(\theta))$ where $\theta = \mathrm{vec}[\mathbf{w}^{(1),\mathrm{T}}, \ldots, \mathbf{w}^{(1),\mathrm{T}}]^\mathrm{T}$ is the stacked vector of weights in all n layers. The Hessian matrix can be computed by

$$\frac{\partial^2 L}{\partial \theta_i \partial \theta_j} = \frac{\partial}{\partial \theta_j}\left(\sum_{k=1}^m \frac{\partial L}{\partial \mathbf{z}_k^{(n)}} \frac{\partial \mathbf{z}_k^{(n)}}{\partial \theta_i}\right) = \sum_{k=1}^m \frac{\partial L}{\partial \mathbf{z}_k^{(n)}} \frac{\partial^2 \mathbf{z}_k^{(n)}}{\partial \theta_i \partial \theta_j} + \sum_{k,l=1}^m \frac{\partial \mathbf{z}_k^{(n)}}{\partial \theta_i} \frac{\partial^2 L}{\partial \mathbf{z}_k^{(n)} \partial \mathbf{z}_l^{(n)}} \frac{\partial \mathbf{z}_l^{(n)}}{\partial \theta_j}. \tag{96}$$

Since the pretrained model has fully converged by default, the gradient term $\bar{\mathbf{g}}^{(\mathbf{w})}$ in Eqn. (95) and $\nabla_{\mathbf{z}^{(n)}} L$ in Eqn. (96) tend to zero, and the Hessian matrix can be approximated as:

$$\mathbf{H}^{(\theta)} \approx \mathbf{J}_{\mathbf{z}^{(n)}}(\theta)^\mathrm{T} \mathbf{H}^{(\mathbf{z}^{(n)})} \mathbf{J}_{\mathbf{z}^{(n)}}(\theta), \tag{97}$$

where $\mathbf{J}_{\mathbf{z}^{(n)}}(\theta)$ is the Jacobian matrix of the network output with respect to the network parameters. By approximating the product of the perturbation and the Jacobian as the first-order Taylor expansion of the network output perturbation $\Delta\mathbf{z}^{(n)}$, the objective is transformed from minimizing Eqn. (95) to minimizing $\Delta\mathbf{z}^{(n)}$:

$$\Delta\mathbf{z}^{(n)} = \hat{\mathbf{z}}^{(n)} - \mathbf{z}^{(n)} \approx \mathbf{J}_{\mathbf{z}^{(n)}}(\theta)\Delta\theta. \tag{98}$$

$$\arg\min_{\hat{\theta}} \Delta\theta^\mathrm{T} \bar{\mathbf{H}}^{(\theta)} \Delta\theta \approx \arg\min_{\hat{\theta}} \mathbb{E}\left[\Delta\mathbf{z}^{(n),\mathrm{T}} \mathbf{H}^{(\mathbf{z}^{(n)})} \Delta\mathbf{z}^{(n)}\right]. \tag{99}$$

It is important to note that the necessary condition for applying a Taylor expansion is that $\Delta\mathbf{x}$ tends to zero. However, under low-bit quantization, both weight and activation perturbations are sufficiently large, causing the conditions in Eqn. (98) and Eqn. (99) to be violated. As a result, the alignment objective between the full-precision module and the quantized module falls outside the limited optimization range of the quantization parameters, making convergence difficult. Therefore, [55] further demonstrates the necessity of parameter fine-tuning under low-bit quantization.

**Quantization noise also has a certain beneficial effect.** Additionally, we observe an intriguing phenomenon: some methods perform even better under W4A8 quantization than under W8A8. For example, in the W4A8 quantization setting, D2-DPM [46] shows an increase of 24.25 in IS and a reduction of 1.32 in sFID, whereas EfficientDM [57] and QuEST [57] exhibit a decrease of 2.75 and 2.41 in FID, and a decrease of 2.27 and 1.21 in sFID, respectively. To explain this, we draw upon previous theoretical research. Karras et al. [144] show that SDE can be generalized as a combination of the probability flow ODE and a time-varying Langevin diffusion SDE:

$$\mathrm{d}\mathbf{x}_\pm = \underbrace{-\dot{\sigma}(t)\sigma(t)\nabla_\mathbf{x} \log p(\mathbf{x}; \sigma(t))\,\mathrm{d}t}_{\text{probability flow ODE}} \pm \underbrace{\beta(t)\sigma(t)^2 \nabla_\mathbf{x} \log p(\mathbf{x}; \sigma(t))\,\mathrm{d}t + \sqrt{2\beta(t)}\sigma(t)\,\mathrm{d}\omega}_{\text{Langevin diffusion SDE}}, \tag{100}$$

Moreover, stochasticity in SDE effectively contracts accumulated error, which consists of both the discretization error along the trajectories as well as the approximation error of the learned neural network relative to the ground truth drift [145]. Therefore, we hypothesize that the quantization noise with an appropriate intensity introduced under W4A8 creates additional variance, which further compensates for the stochasticity, leading to a superior Langevin SDE compared to the original SDE (ODE), and consequently results in a positive performance gain.

### 4.2.2 Unconditional Image Generation

Table. 3 presents the performance evaluation results of various methods on the unconditional image generation task across two benchmarks: LSUN-Bedrooms (LDM-4) and LSUN-Churches (LDM-8).

Table 3: Performance evaluation of various approaches for unconditional image generation, conducted using LDM-4 ($\eta = 1.0, steps = 200$) on the LSUN-Bedrooms 256×256 dataset and LDM-8 ($\eta = 0.0, steps = 200$) on the LSUN-Churches 256×256 dataset.

| Method | Bits (W/A) | LSUN-Bedrooms 256 × 256 | | | | LSUN-Churches 256 × 256 | | | |
|---|---|---|---|---|---|---|---|---|---|
| | | FID ↓ | sFID ↓ | Precision ↓ | Recall ↓ | FID ↓ | sFID ↓ | Precision ↑ | Recall ↑ |
| FP | 32/32 | 3.03 | 7.03 | 64.65% | 47.60% | 4.17 | 12.91 | 66.00% | 51.46% |
| PTQ4DM [40] | 8/8 | 12.97 | 17.50 | 52.26% | 40.17% | 17.44 | 18.69 | 49.16% | 48.00% |
| Q-Diffusion [41] | 8/8 | 7.37 | 12.79 | 53.62% | 41.73% | 8.97 | 13.81 | 53.42% | 51.04% |
| EDA-DM [50] | 8/8 | 7.23 | 12.28 | 55.36% | 42.88% | 8.12 | 12.32 | 56.14% | 52.58% |
| TFMQ-DM [52] | 8/8 | 7.62 | 11.98 | 54.59% | 42.10% | 8.77 | 13.09 | 55.62% | 51.38% |
| QNCD [107] | 8/8 | 9.84 | 13.50 | 51.27% | 45.34% | 9.10 | 14.71 | 53.29% | 50.39% |
| PTQD [44] | 8/8 | 9.16 | 12.94 | 51.99% | 44.32% | 8.31 | 12.97 | 56.57% | 54.15% |
| D$^2$-DPM [46] | 8/8 | 7.55 | 12.56 | 55.60% | 45.80% | 7.83 | 12.51 | 56.86% | **54.54%** |
| EfficientDM[†] [57] | 8/8 | **4.39** | **9.02** | **60.68%** | 45.72% | **6.89** | **12.13** | **61.92%** | 51.77% |
| QuEST[†] [55] | 8/8 | 4.52 | 9.31 | 60.11% | **46.20%** | 7.14 | 13.29 | 57.43% | 52.83% |
| PTQ4DM [40] | 4/8 | 15.12 | 20.97 | 46.96% | 37.41% | 22.58 | 23.87 | 44.63% | 44.09% |
| Q-Diffusion [41] | 4/8 | 11.24 | 16.70 | 52.67% | 40.82% | 13.57 | 15.83 | 51.29% | 48.71% |
| EDA-DM [50] | 4/8 | 10.61 | 14.36 | 52.35% | 39.43% | 13.28 | 16.29 | 51.86% | 47.53% |
| TFMQ-DM [52] | 4/8 | 10.96 | 15.81 | 52.59% | 40.20% | 12.94 | 16.19 | 54.70% | 50.26% |
| QNCD [107] | 4/8 | 13.26 | 16.59 | 49.54% | 41.62% | 14.10 | 17.59 | 48.73% | 47.15% |
| PTQD [44] | 4/8 | 12.57 | 16.04 | 51.31% | 42.40% | 12.96 | 15.42 | 50.23% | 52.80% |
| D$^2$-DPM [46] | 4/8 | 10.72 | 15.23 | 51.44% | 43.90% | 11.18 | 15.14 | 52.27% | **53.78%** |
| EfficientDM[†] [57] | 4/8 | 6.97 | 10.39 | 54.72% | 42.94% | **8.36** | 14.79 | 54.61% | 49.72% |
| QuEST[†] [55] | 4/8 | **6.54** | **10.06** | **55.42%** | 43.18% | 8.93 | 13.84 | **55.18%** | 50.33% |
| EfficientDM[†] [57] | 4/4 | 10.79 | 14.10 | **46.76%** | 32.53% | 15.32 | 20.57 | 41.87% | **29.84%** |
| QuEST[†] [55] | 4/4 | **8.92** | **13.21** | 46.44% | **34.59%** | **12.54** | **18.30** | **42.06%** | 27.40% |

Here, we still categorize the methods into three subgroups for analysis: 1) Calibration customization (PTQ4DM [40], Q-Diffusion [41], and EDA-EDA-QDM [50]); 2) Quantization error correction (QNCD [107], PTQD [44], D$^2$-DPM [46]); 3) Quantization-aware training (EfficientDM [57], QuEST [55]).

**Calibration customization.** In the unconditional generation task, compared to normal distribution time-step sampling [40] and uniform time-step sampling [41], the enhanced distribution-aligned time-step sampling [50] performs better on the overall metric. Under both W8A8 and W4A8, it further reduces the FID/sFID average by 0.495/1.0 and 0.46/0.94 compared to the second-best method [41].

**Quantization error correction.** In the unconditional generation task, PTQD [44] and D$^2$-DPM [46] outperform QNCD in overall metrics. Additionally, all error correction algorithms are closer to the full-precision model on the LSUN-Churches (LDM-8) benchmark than on the LSUN-Bedrooms (LDM-4) benchmark. We believe this is closely related to variance overflow during the sampling process. The former uses DDIM sampling with $\eta = 0.0$, while the latter sets $\eta = 1.0$, effectively degrading to DDPM sampling. The larger native variance limits the SDE's ability to absorb additional accumulated errors. This can be intuitively understood as the full-precision trajectory variance being already large, and after quantization, the variance is more prone to explosion. Therefore, we recommend setting a smaller $\eta$ during quantization. **Quantization-aware training.** In the unconditional generation task, the weight-tuning-based EfficientDM [57] and QuEST [55] achieve a clear lead in overall metrics, with an average increase of only FID by 4.07 and 4.14 on the full-precision model under W4A8. Even on the LSUN-Bedrooms (LDM-4) benchmark, where full-precision sampling variance is relatively large, they still outperform the optimal PTQ-based method in FID/sFID with an average lead of 2.56/1.87 and 2.43/1.83 , without exhibiting significant variance explosion. This suggests that weight tuning effectively reduces the additional variance introduced.

Table 4: The performance comparison of text-guided image generation with Stable-Diffusion v1-4 [25] on MS-COCO [91] captions.

| Method | Bits (W/A) | Size(MB) | BOPs(T) | FID ↓ | sFID ↓ | CLIP-score ↑ |
|---|---|---|---|---|---|---|
| FP | 32/32 | 3438 | 753 | 13.27 | 19.46 | 0.3064 |
| Q-Diffusion [41] | 8/8 | 871 | 47.05 | 13.28 | 20.65 | 0.2904 |
| TFMQ-DM [52] | 8/8 | 871 | 47.05 | 14.62 | 23.05 | 0.2852 |
| PCR [111] | 8/8 | 871 | 47.05 | 25.91 | 32.49 | 0.2637 |
| DGQ [110] | 8/8 | 871 | 47.05 | **13.20** | **20.17** | **0.2968** |
| Q-Diffusion [41] | 4/8 | 436 | 23.53 | 14.40 | **21.09** | 0.2875 |
| TFMQ-DM [52] | 4/8 | 436 | 23.53 | 14.97 | 24.06 | 0.2821 |
| PCR [111] | 4/8 | 436 | 23.53 | 24.15 | 30.80 | 0.2623 |
| DGQ [110] | 4/8 | 436 | 23.53 | **13.92** | 22.16 | **0.2904** |

### 4.2.3 Text-guided Image Generation

Table. 4 presents the performance evaluation results of four representative open-source solutions for text-guided image generation. Both Q-Diffusion [41] and TFMQ-DM [52] perform well on overall metrics, indicating that applying split quantization mitigates the impact of the bimodal data distribution, and that temporal feature alignment remains effective for the quantization of text-to-image models. However, the overall best-performing DGQ [110] maintains a clear advantage under both W8A8 and W4A8 settings, particularly with the higher CLIP-score, which indicates its superior ability to preserve text-image consistency.

## 4.3 Qualitative Analysis

Initially, a categorical summary of the interference effects introduced by quantization is provided based on the visualization results. Then, from the perspective of trajectory analysis, we explain how quantization noise interferes with the sampling process, ultimately leading to visualization distortions.

### 4.3.1 Visualization Results Analysis

Fig. 3 presents a visual comparison of the LDM-4 quantized using each scheme on the ImageNet 256×256 dataset. It is evident that the interference effects introduced by quantization noise in the generated images can be attributed to color bias, pixel overexposure, alterations in structural features, and blurring of details.

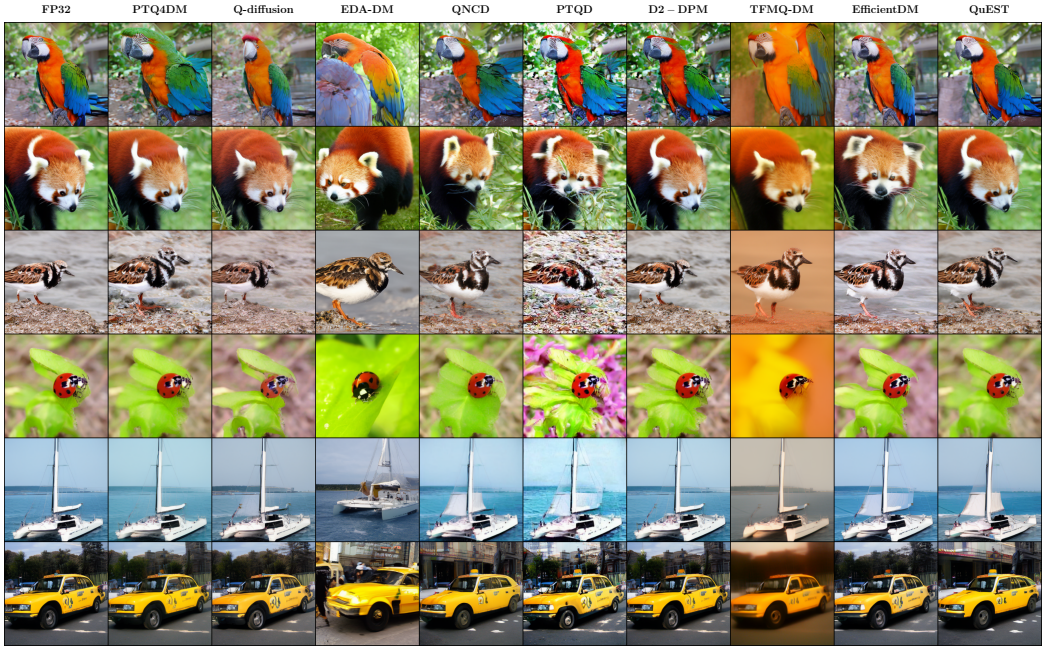

Figure 3: Visual comparison of the full-precision LDM-4 and its W4A8 quantized versions using PTQ4DM [40], Q-Diffusion [41], EDA-DM [50], QNCD [107], PTQD [44], $D^2$-DPM [46], TFMQ-DM [52], EfficientDM [57], and QuEST [55] on the ImageNet 256×256 dataset.

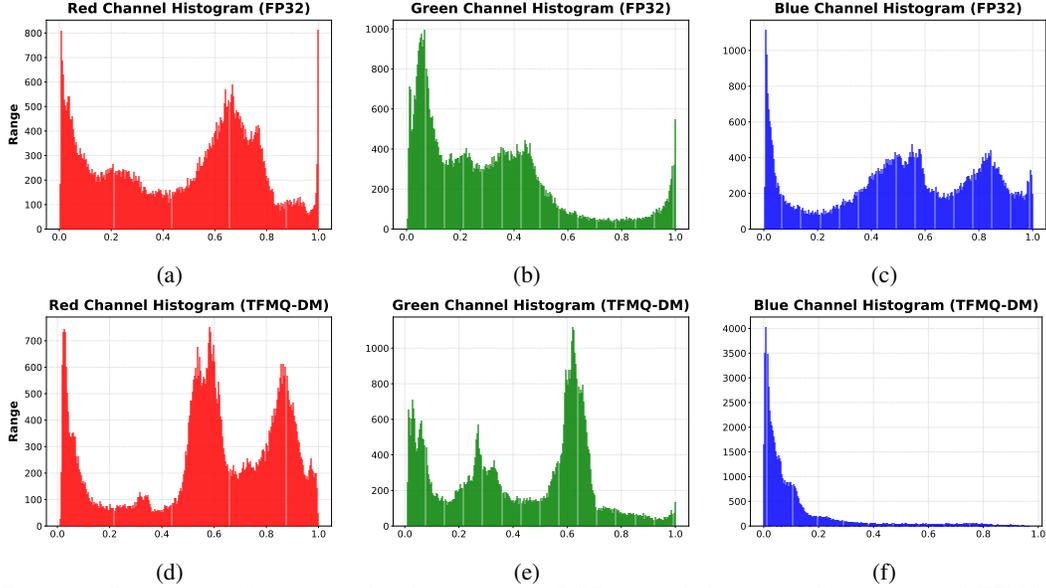

Figure 4: Comparison of activation distributions in the RGB channels between full-precision and TFMQ-DM [52] under W4A8 quantization. (a) Activation distribution of the red channel in FP32 precision. (b) Activation distribution of the green channel in FP32 precision. (c) Activation distribution of the blue channel in FP32 precision. (d) Activation distribution of the red channel under TFMQ-DM with W4A8 quantization. (e) Activation distribution of the green channel under TFMQ-DM with W4A8 quantization. (f) Activation distribution of the blue channel under TFMQ-DM with W4A8 quantization.

**Color Bias.** The uneven distribution of quantization noise across different color channels leads to color drift. As shown in Fig. 3, under W4A8, the image generated by TFMQ-DM [52] exhibits an overall shift towards a dark yellow domain, presenting a more noticeable color bias compared to other schemes. To further analyze this, Fig. 4 illustrates a comparison of the activation distributions between the full-precision and quantized activations, with a specific focus on the differences across the RGB channels. It is clearly observed that quantization noise significantly alters the activation distribution of the blue channel, resulting in a more pronounced mean shift compared to the red and green channels.

**Pixel Overexposure.** The overlap of high variance activations across channels directly results in pixel overexposure. This amplification of variance exhibits a dual nature. As demonstrated in Fig. 3, the PTQD [44] results under W4A8 quantization display a stronger overall image contrast compared to the full-precision model and most other schemes. In certain samples, the moderate accumulation of high variance enhances the image contrast, improving visual quality. However, in other samples, excessive variance introduces pixel artifacts, manifesting as noise. Further analysis in Fig. 5 attributes this high variance phenomenon to the global quantization noise interference, the increased occurrence of extreme values caused by quantization noise and the consistent additive shift observed across the RGB channels. In addition, analysis of the sampling trajectories further reinforces the presence of high variance in PTQD [44].

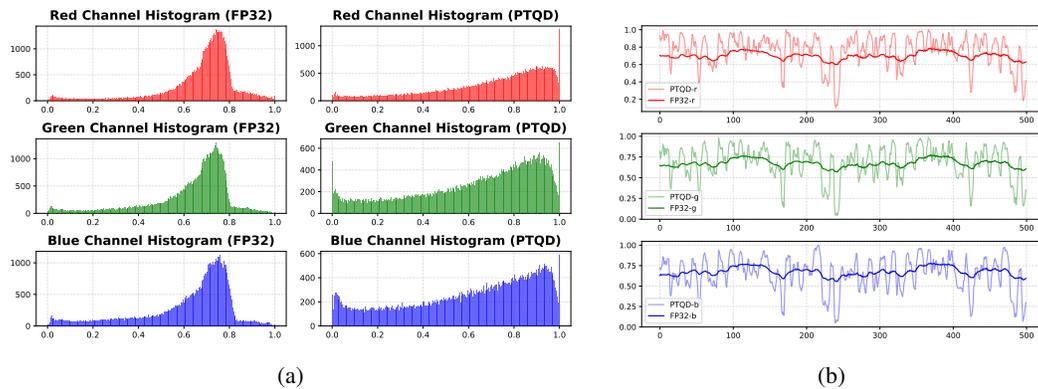

Figure 5: Comparison of activation distributions in the RGB channels between full-precision and PTQD [44] under W4A8 quantization. (a) Quantization noise significantly affects the entire pixel domain in the W4A8 PTQD quantized model, causing an increase in the frequency of extreme values. Additionally, the noise interference patterns converge across all three channels. (b) Activation variations before and after quantization across the RGB channels for each element. The noise interference across channels for each element exhibits a significant cumulative effect.

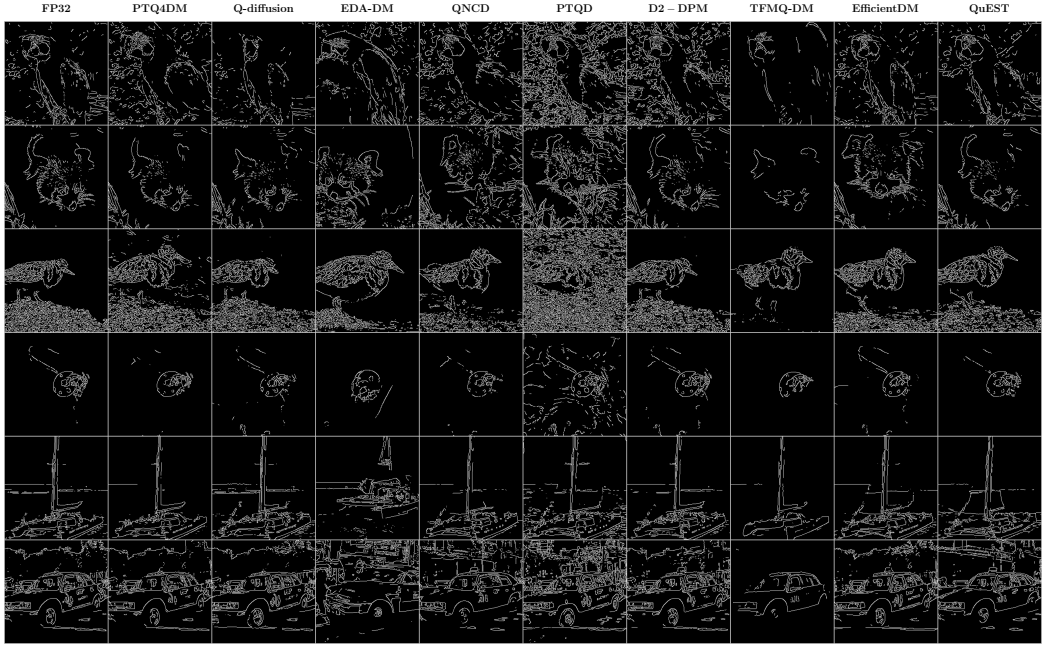

Figure 6: Visual comparison of the edge detection results of the full-precision LDM-4 and its quantized versions using PTQ4DM [40], Q-Diffusion [41], EDA-DM [50], QNCD [107], PTQD [44], $D^2$-DPM [46], TFMQ-DM [52], EfficientDM [57], and QuEST [55] on the ImageNet 256×256 dataset.

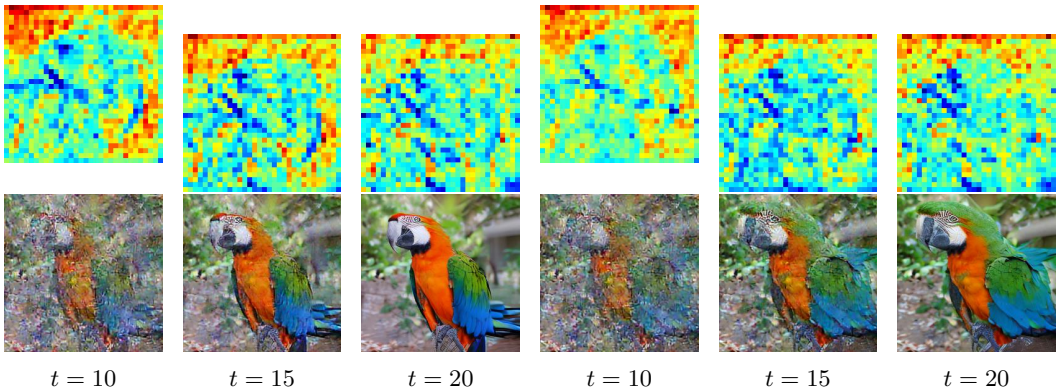

$t = 10$     $t = 15$     $t = 20$     $t = 10$     $t = 15$     $t = 20$

Figure 7: The comparison of cross-attention before and after quantization in the LDM4 model using DDIM for 20-step sampling, with intermediate steps shown (with $t = 0$ representing the denoising starting point, where the samples are pure noise). The first row visualizes heatmaps of the attention weights of $model.output_blocks.8.1.transformer_blocks.0.attn2$ corresponding to category-conditioned tokens. The second row illustrates the generated samples at intermediate steps of the denoising process. The first three columns correspond to the full-precision model, while the last three columns display the results with PTQ4DM [40] quantization (W4A8).

**Loss of Subtle Details.** Quantization maps continuous full-precision activations to a discrete set of regions, leading to activation shifts within the color gamut corresponding to specific frequency bands, which visually manifests as a loss of image details. As shown in Fig. 6, Canny edge detection results demonstrate that QAT-based methods, such as EfficientDM [57] and QuEST [55], effectively preserve the details of the generated subject. In the case of PTQ-based methods, TFMQ-DM [52] loses significant details due to noticeable color bias caused by distribution shifts. PTQ4DM [40], while retaining most of the details, loses some high-frequency information, particularly in the hair region. Q-Diffusion [41] slightly loses some texture details. The variance introduced by PTQD [44] enhances certain details, but also leads to some interference. $D^2$-DPM introduces minimal variance in the background while largely preserving the subject's details.

**Alterations in the Structural Characteristics.** In conditional generation tasks, cross-attention is typically employed to fuse image features and cross-modal conditional features, thereby imposing control constraints on image reconstruction. However, the attention weight shifts caused by quantization disturbances interfere with the conditioning control over image generation, resulting in alterations to the structured features of the image in the spatial domain. As shown in Fig. 7, the heatmaps of cross-attention weights for full precision (left three columns) and those quantized using PTQ4DM with W4A8 (right three columns) indicate that the quantized attention is interfered with by quantization noise, which affects the boundary judgment of the generated object (i.e., the parrot). This interference leads to a potential inclusion of part of the background into the generated object, ultimately causing an increase in the area of structural features—specifically, the wings. It is particularly noteworthy that the impact mainly stems from quantization disturbances early in the denoising process (when $t \leq T/2$), leading to unclear boundary delineation. This reflects the fact that during the earlier stages, the model focuses more on capturing structural features rather than fine details.

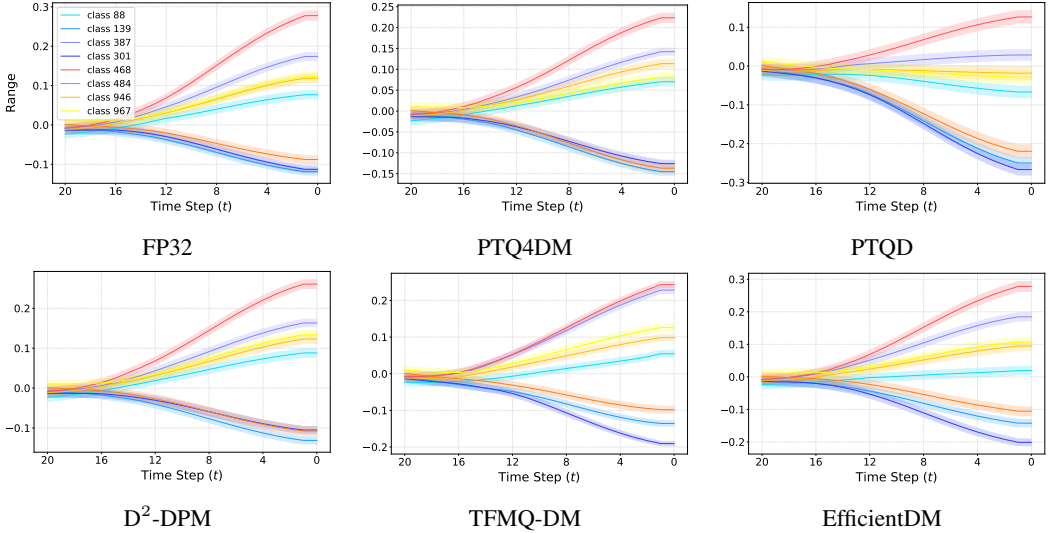

Figure 8: The DDIM 20-step sampling trajectories of eight sample categories from the full-precision LDM-4 model and its quantized versions, including PTQ4DM [40], PTQD [44], $D^2$-DPM [46], TFMQ-DM [52], and EfficientDM [57], are shown on the ImageNet256×256 dataset.

#### 4.3.2 Trajectory Analysis

In Fig. 8, we visualize and compare several representative sampling trajectories. It can be observed that different sampling methods exhibit varying degrees of distribution drift and variance convergence. PTQ4DM [40] exhibits consistent distribution drift across most categories, with varying degrees of drift observed among different categories, particularly in class 484. However, the data variance is approximately equivalent to that of the full-precision activations. PTQD [44] effectively controls the relative convergence of sampling trajectories across categories, suggesting that the quantized model retains good class-conditioned perception. However, it exhibits a slight global systematic shift, accompanied by a slightly higher sample variance. $D^2$-DPM [46] aligns most of the category sample trajectories with those of the full-precision model, maintaining better relative convergence across categories compared to PTQ4DM [40], particularly for the 946 and 967 class.

The sampling trajectories of TFMQ-DM [52] in categories 301 and 387 exhibit clear deviations, which we attribute to significant asynchronous drift in the activation values between channels, leading to the chromatic aberration phenomenon discussed earlier. EfficientDM [57] also effectively controls the relative convergence across categories, demonstrating stronger class discriminability compared to other methods. This may be due to the weight fine-tuning, which enhances feature representation under quantization-induced perturbations. However, this approach may result in slightly lower alignment of specific features at the sample level compared to other methods, as evidenced by the fact that some features, while not perfectly aligned with the full-precision model in Fig. 3, still remain reasonable.

## 5 Future Prospects

This section provides a brief discussion of future research directions that can facilitate and guide the development of diffusion model quantization.

**Integration of Quantization with Advanced Training Strategie.** Further research could focus on integrating quantization techniques with emerging training methodologies, such as self-supervised learning and few-shot learning, to improve the robustness and efficiency of diffusion models in low-resource settings.

**Quantization-Aware Fine-Tuning for Enhanced Performance.** Investigating the combination of quantization with fine-tuning approaches could lead to methods that are both memory-efficient and performance-preserving, ensuring that diffusion models remain competitive even after significant quantization.

**Optimizing Vector Quantization in Diffusion Models.** While much of the research has concentrated on quantizing weights and activations, future studies could investigate the potential of vector quantization techniques to improve the efficiency of diffusion models. Specifically, applying vector quantization to the discrete representation of latent variables or embeddings could lead to more compact and computationally efficient models, facilitating improved performance across various modalities, including image, audio, and text.

**Adapting Quantization for Real-World Deployment.** Research could be directed towards adapting quantization techniques to specific hardware architectures, such as edge devices and specialized AI accelerators, to ensure that diffusion models are both effective and efficient when deployed in real-world applications with limited resources.

**Exploring the Trade-offs Between Quantization and Model Interpretability.** Investigating how quantization impacts the interpretability and explainability of diffusion models could provide valuable insights into the balance between model efficiency and the ability to understand and trust model decisions, especially in critical applications.